\documentclass{article}

\usepackage{arxiv}

\usepackage[utf8]{inputenc} 
\usepackage[T1]{fontenc}    
\usepackage{url}            
\usepackage{booktabs}       

\usepackage{amsmath,amssymb,amsfonts,amsthm,dsfont,mathrsfs,upgreek}
\usepackage{latexsym,breqn}
\usepackage{nicefrac}       
\usepackage{microtype}      
\usepackage{natbib}

\usepackage{graphicx,subfigure}

\usepackage{algorithm}
\usepackage[noend]{algpseudocode}
\makeatletter
\def\BState{\State\hskip-\ALG@thistlm}

\usepackage{ctable,multicol,multirow}

\usepackage{color}
\definecolor{green}{rgb}{0, 0.5, 0}
\usepackage[pdftex,
			bookmarks, bookmarksnumbered=true, bookmarksopen=true,
			colorlinks, citecolor=blue, filecolor=black, linkcolor=black, urlcolor=green,
			pdfauthor={López-Lopera, Andrés Felipe}, pdftitle={PhDEMSE},
			pdftoolbar=true, pdfmenubar=true, pdffitwindow = true
			]{hyperref}

\newcommand{\realset}[1]{\mathds{R}^{#1}}

\newcommand{\normrnd}[2]{\mathcal{N}\left({#1,#2}\right)}

\newcommand{\BA}{\boldsymbol{A}}
\newcommand{\Bc}{\boldsymbol{c}}
\newcommand{\Bl}{\boldsymbol{l}}
\newcommand{\Bu}{\boldsymbol{u}}

\newcommand{\Bs}{\textbf{s}}
\newcommand{\Bx}{\textbf{x}} 

\newcommand{\Bxi}{\boldsymbol{\xi}}

\newcommand{\Bchi}{\boldsymbol{\chi}}
\newcommand{\BGamma}{\boldsymbol{\Gamma}}
 
\newcommand{\Bphi}{\boldsymbol{\phi}} 
\newcommand{\Bzero}{\boldsymbol{0}} 
 \newcommand{\BSigma}{\boldsymbol{\Sigma}}
\newcommand{\Bmu}{\boldsymbol{\mu}}
\newcommand{\Btheta}{\boldsymbol{\theta}} 
\newcommand{\dee}{\mathrm{d}}

\title{Gaussian Process Modulated Cox Processes under Linear Inequality Constraints}
\date{}
\author{
  Andr\'es F. L\'opez-Lopera\thanks{Part of this work was completed during an internship of A. F. L\'opez-Lopera at PROWLER.io.} \\
  Mines Saint-\'Etienne\\
  42000 Saint-\'Etienne, France \\
  \texttt{andres-felipe.lopez@emse.fr} \\
  \And
  ST John \\
  PROWLER.io\\
  Cambridge, CB2 1LA, UK \\
  \texttt{st@prowler.io} \\
  \And
  Nicolas Durrande\\
  PROWLER.io \\
  Cambridge, CB2 1LA, UK \\
  \texttt{nicolas@prowler.io} \\
}

\begin{document}
\maketitle

\begin{abstract}
	Gaussian process (GP) modulated Cox processes are widely used to model point patterns. Existing approaches require a mapping (link function) between the unconstrained GP and the positive intensity function. This commonly yields solutions that do not have a closed form or that are restricted to specific covariance functions. We introduce a novel finite approximation of GP-modulated Cox processes where positiveness conditions can be imposed directly on the GP, with no restrictions on the covariance function. Our approach can also ensure other types of inequality constraints (e.g.\ monotonicity, convexity), resulting in more versatile models that can be used for other classes of point processes (e.g.\ renewal processes). We demonstrate on both synthetic and real-world data that our framework accurately infers the intensity functions. Where monotonicity is a feature of the process, our ability to include this in the inference improves results.
\end{abstract}


\section{Introduction}
Point processes are used in a variety of real-world problems for modelling temporal or spatiotemporal point patterns in fields such as astronomy, geography, and ecology \citep{Baddeley2015SpatStatsR,Moller2004SpatialPointProcess}. In reliability analysis, they are used as renewal processes to model the lifetime of items or failure (hazard) rates \citep{Cha2018SpatPointReliability}. 

Poisson processes are the foundation for modelling point patterns \citep{Kingman1992PoissonProcess}. Their extension to stochastic intensity functions, known as {doubly stochastic Poisson processes} or Cox processes \citep{Cox1955}, enables non-parametric inference on the intensity function and allows expressing uncertainties \citep{Moller2004SpatialPointProcess}. Moreover, previous studies have shown that other classes of point processes may also be seen as Cox processes. For example, \citet{Yannaros1988GamaProcesses} proved that Gamma renewal processes are Cox processes under non-increasing conditions. A similar analysis was made later for Weibull processes \citep{Yannaros1994WeibullProcess}.

Gaussian processes (GPs) form a flexible prior over functions, and are widely used to model the intensity process $\Uplambda(\cdot)$ \citep{Moller2001LogGaussianCPs,Adams2009CPs,Yee2011GPRPs,Gunter2014CPSigmoid,Lasko2014GPGammaProcesses,Lloyd2015CPs,Fernandez2016GPSurvival,Donner2018SigmoidalCPs}. However, to ensure positive intensities, this commonly requires link functions between the intensity process and the GP $g(\cdot)$. Typical examples of mappings are $\Uplambda(x) = \exp(g(x))$ \citep{Moller2001LogGaussianCPs,Diggle2013CPGeo,Flaxman2016CPKronecker} or $\Uplambda(x) = g(x)^2$ \citep{Lloyd2015CPs,Kozachenko2016SimCPs}. The exponential transformation has the drawback that there is no closed-form expression for some of the integrals required to compute the likelihood. Although the square inverse link function allows closed-form expressions for certain kernels, it leads to models exhibiting ``nodal lines'' with zero intensity due to the non-monotonicity of the transformation \citep[see][for a discussion]{ST2018CPs}. Furthermore, current approaches to Cox process inference cannot be used in applications such as renewal processes that require both positivity and monotonicity constraints.

Here, we introduce a novel approximation of GP-modulated Cox processes that does not rely on a mapping to obtain the intensity. In our approach we impose the constraints (e.g.\ non-negativeness or monotonicity) directly on $\Uplambda(\cdot)$ by sampling from a truncated Gaussian vector. This has the advantage that the likelihood can be computed in closed form. Moreover, our approach can ensure any type of linear inequality constraint everywhere, which allows modelling of a broader range of point processes.

This paper is organised as follows. In Section \ref{sec:CoxProcess}, we briefly describe inhomogeneous Poisson processes and some of their extensions. In Sections \ref{sec:finiteCoxProcess} and \ref{sec:CPinference}, we introduce a finite representation of GP-modulated Cox processes and the corresponding Cox process inference under inequality constraints. In Section \ref{sec:infResults}, we apply our framework to 1D and 2D inference examples under different inequality conditions. We also test its performance in reliability applications with hazard rates exhibiting monotonic behaviours. Finally, in Section \ref{sec:conclusions}, we summarise our results and outline potential future work.

	\section{POISSON POINT PROCESSES}
	\label{sec:CoxProcess}
	A Poisson process $X$ is a random countable subset of $\mathcal{S} \subseteq \realset{d}$ where points occur independently \citep{Baddeley2006SpatPointProcess}. Let $N \in \mathds{N}$ be a random variable (r.v.)\ denoting the number of points in $X$. Let $X_1, \cdots, X_n$ be a set of $n$ independent and identically distributed (i.i.d.)\ r.v.'s on $\mathcal{S}$. The likelihood of $(N=n, X_1 = \Bx_1, \cdots, X_n = \Bx_n)$ under an inhomogeneous Poisson process with non-negative intensity $\lambda(\cdot)$ is given by \citep{Moller2004SpatialPointProcess}
	\begin{equation}
	f_{(N, X_1, \cdots, X_n)}(n, \Bx_1, \cdots, \Bx_n)
	= \frac{\exp(-\mu)}{n!} \prod_{i=1}^{n} \lambda(\Bx_i),
	\label{eq:uncondJointPoissonLikelihood}
	\end{equation}
	where
	\begin{equation}
	\mu = \int_{\mathcal{S}} \lambda(\Bs) \,\dee\Bs
	\label{eq:intensitymeasure}
	\end{equation}
	is the {intensity measure} or overall intensity.
	
	When $\mathcal{S}$ is the real line, the distance (inter-arrival time) between consecutive points of a Poisson process follows an exponential distribution.
	Renewal processes are a generalisation of Poisson processes where inter-arrival times are i.i.d.\ but not necessarily exponentially distributed. An example is the Weibull process where inter-arrival times are distributed following $\lambda(x) = \alpha \beta x^{\beta-1}$ \citep{Cha2018SpatPointReliability}.
	
	Cox processes \citep{Cox1955} are a natural extension of inhomogeneous Poisson processes where $\lambda(\cdot)$ is sampled from a non-negative stochastic process $\Uplambda(\cdot)$. Previous studies have shown that many classes of point processes can be seen as Cox processes under certain conditions \citep{Moller2004SpatialPointProcess,Yannaros1988GamaProcesses,Yannaros1994WeibullProcess}. For example, Weibull renewal processes are Cox processes for $\beta \in (0, 1]$ \citep{Yannaros1994WeibullProcess}. This motivates the construction of GPs with non-negative and monotonic constraints, so that they can be used as intensities $\Uplambda(\cdot)$ of Cox processes.
	
	\section{APPROXIMATION OF GP MODULATED COX PROCESSES}
	\label{sec:finiteCoxProcess}
	In this work, we approximate the intensity $\Uplambda(\cdot)$ of the Cox process by a finite-dimensional GP $\Uplambda_m(\cdot)$ subject to some inequality constraints (e.g.\ boundedness, monotonicity, convexity). Since positiveness constraints are imposed directly on $\Uplambda_m(\cdot)$, a link function is no longer necessary. This has two main advantages. First, the likelihood \eqref{eq:uncondJointPoissonLikelihood} can be computed analytically. Second, as our approach ensures any linear inequality constraint, it can be used for modelling a broader range of point processes.
	\begin{figure*}
		\centering
		\subfigure[\label{subfug:lineqGP1} unconstrained GP]{\includegraphics[width=0.325\textwidth]{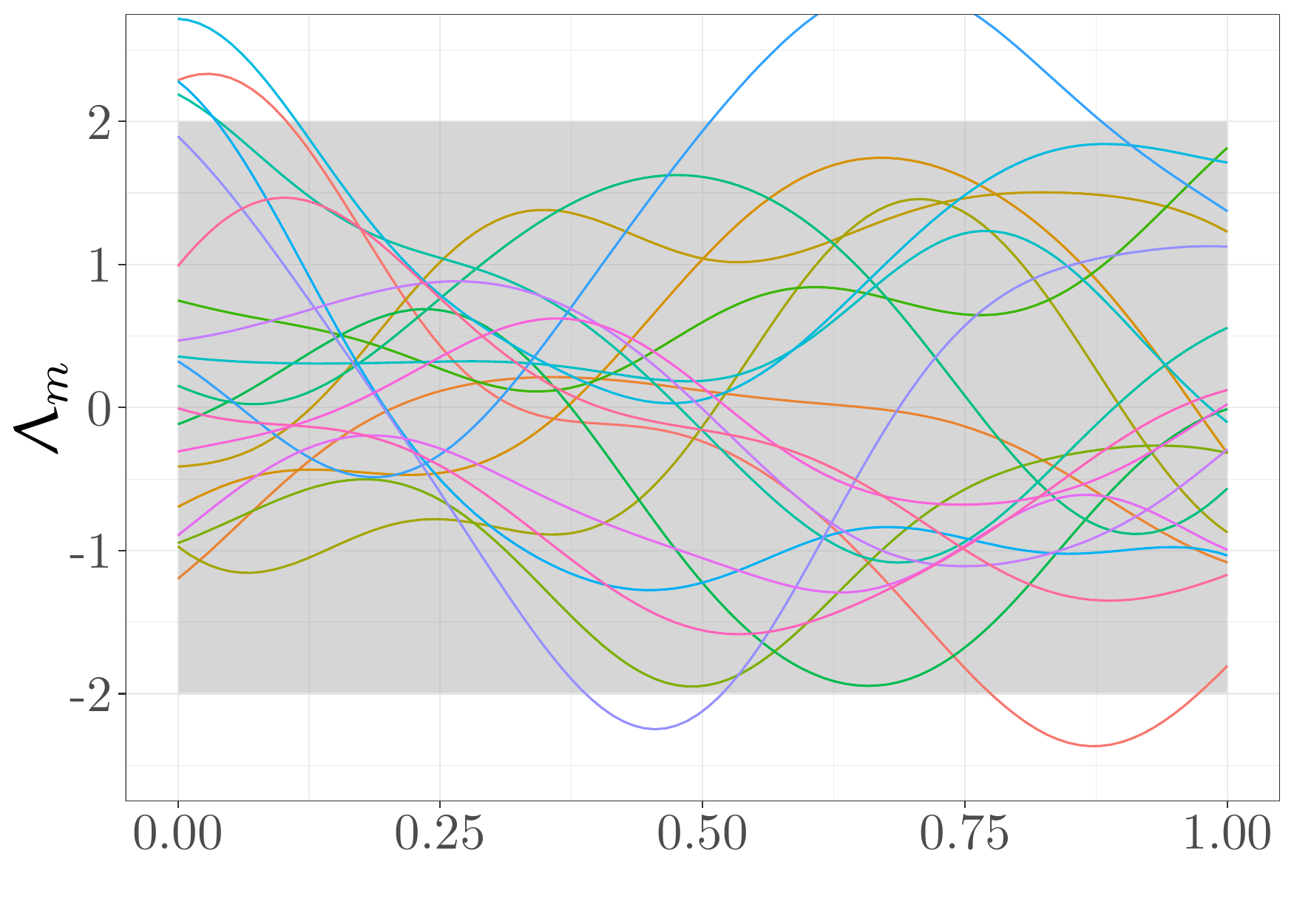}}
		\subfigure[\label{subfug:lineqGP2} GP with $\mathcal{C}_+$ constraints]{\includegraphics[width=0.325\textwidth]{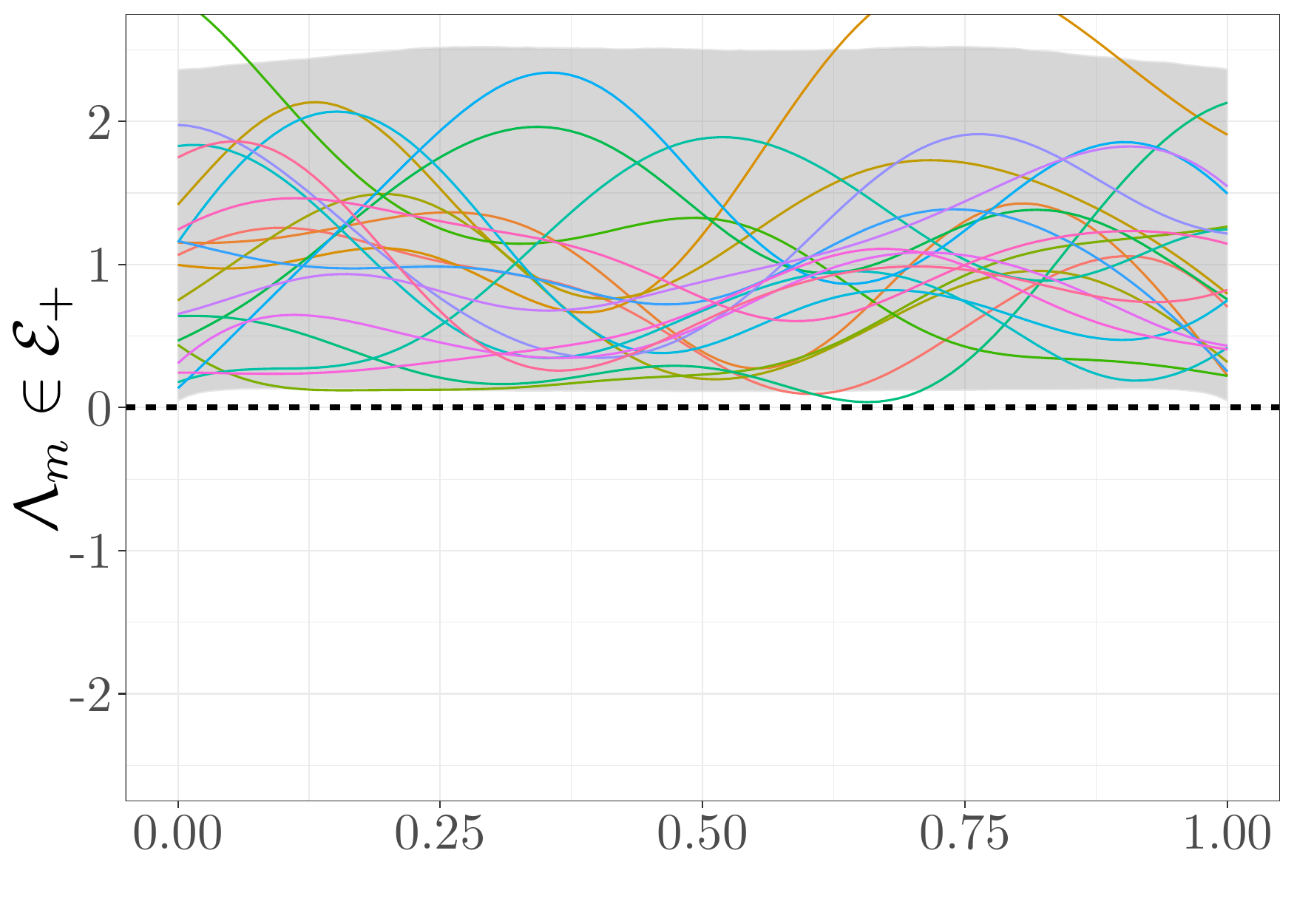}}
		\subfigure[\label{subfug:lineqGP3} GP with $\mathcal{C}_+^\downarrow$ constraints]{\includegraphics[width=0.325\textwidth]{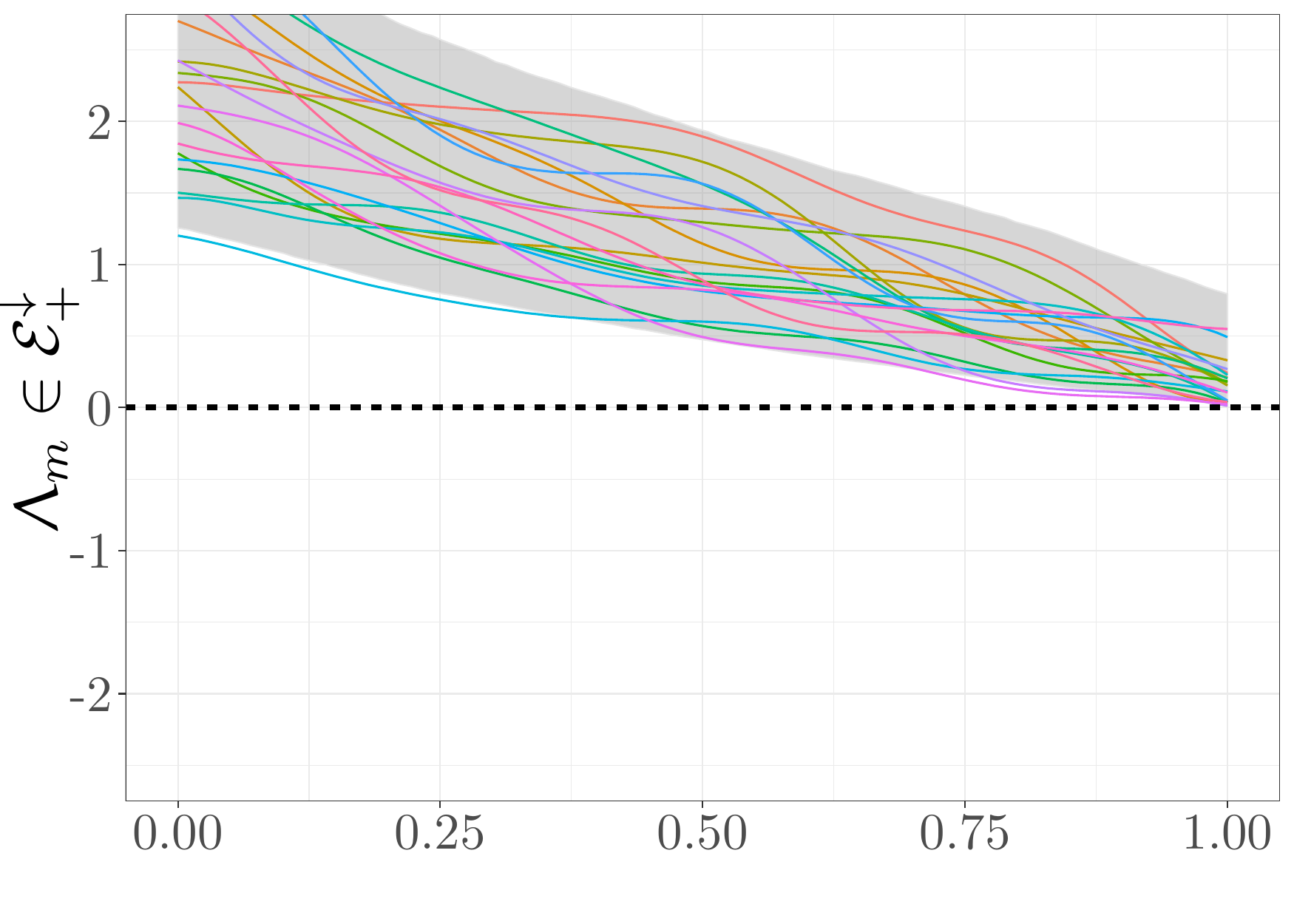}}
		\caption{Samples from the prior $\Uplambda_m(\cdot)$ under (a) no constraints, (b) non-negativeness constraints, (c) both non-negativeness and non-increasing constraints. The grey region shows the $95\%$ confidence interval.}
		\label{fig:lineqGPexamples}
	\end{figure*}
	
	\subsection{Finite Approximation of 1D GPs}
	\label{subsec:finiteGaussianProcess}
	
	Let $\Uplambda(\cdot)$ be a zero-mean GP on $\realset{}$ with arbitrary covariance function $k$. Consider $x \in \mathcal{S}$, with compact space $\mathcal{S} = [0, 1]$, and a set of knots $t_1, \cdots, t_m \in \mathcal{S}$. Here we consider equispaced knots {$t_j = (j-1) \Delta_m$} with $\Delta_m = 1/(m-1)$.	We define $\Uplambda_m(\cdot)$ as the finite-dimensional approximation of $\Uplambda(\cdot)$ consisting of its piecewise-linear interpolation at knots $t_1, \cdots, t_m$, i.e.,
	\begin{equation}
	\Uplambda_m (x) = \sum_{j=1}^{m} \phi_j (x) \xi_j,
	\label{eq:finApprox}
	\end{equation}
	where $\xi_j := \Uplambda(t_j)$ for $j =1, \cdots, m$, and $\phi_1, \cdots, \phi_m$ are hat basis functions given by
	\begin{equation}
	\phi_j (x) :=
	\begin{cases}
	1 - \left|\frac{x - t_j}{\Delta_m}\right| & \mbox{if } \left|\frac{x - t_j}{\Delta_m}\right| \leq 1,\\
	0 & \mbox{otherwise}.
	\end{cases}
	\label{eq:hatbasisfun}
	\end{equation}
	Similarly to spline-based approaches \citep[e.g.,][]{Sleeper1990CPSplines}, we assume that $\Uplambda(\cdot)$ is piecewise defined by (first-order) polynomials. The striking property of this basis is that satisfying the inequality constraints (e.g.\ boundedness, monotonicity, convexity) at the knots implies that the constraints are satisfied everywhere in the input space \citep{Maatouk2017GPineqconst}. Although it is tempting to generalise the above construction to smoother basis functions, it makes this property difficult to enforce.
	
	We aim at computing the distribution of $\Uplambda_m(\cdot)$ under the condition that it belongs to a convex set of functions $\mathcal{E}$ defined by some inequality constraints (e.g.\ positivity). This piecewise-linear representation has the benefit that satisfying $\Uplambda_m(\cdot) \in \mathcal{E}$ is equivalent to satisfying only a finite number of inequality constraints. More precisely,
	\begin{equation}
	\Uplambda_m(\cdot) \in \mathcal{E} \; \Leftrightarrow \; \Bxi \in \mathcal{C},
	\label{eq:convexEqui}
	\end{equation}
	where $\Bxi = [\xi_1, \cdots, \xi_m]^\top$, and $\mathcal{C}$ is a convex set on $\realset{m}$.	For non-negativeness conditions $\mathcal{E}_+$, $\mathcal{C}$ is given by
	\begin{equation}
	\mathcal{C}_+ := \{ c \in \realset{m}; \ \forall \ j = 1, \cdots, m \; : \; c_j \geq 0  \},
	\label{eq:convexsetKnotsPositive}
	\end{equation}
	and for non-increasing conditions $\mathcal{E}_{\downarrow}$, $\mathcal{C}$ is given by
	\begin{equation}
	\mathcal{C}_{\downarrow}:= \{ c \in \realset{m}; \ \forall \ j = 2, \cdots, m \; : \; c_{j-1} \geq c_{j}\}.
	\label{eq:convexsetKnotsMonotone}
	\end{equation}
	Constraints can be composed, e.g.\ the convex set of non-negativeness and non-increasing conditions is given by $\mathcal{C}_+^{\downarrow} = \mathcal{C}_{+}  \cap \mathcal{C}_{\downarrow}$.
	
	Assuming that $\Bxi$ is zero-mean Gaussian-distributed with covariance matrix $\BGamma= (k(t_i,t_j))_{1 \leq i,j \leq m}$, then the distribution of $\Bxi$ conditioned on $\Bxi \in \mathcal{C}$ is a truncated Gaussian distribution. Then, quantifying uncertainty on $\Uplambda_m$ relies on sampling $\Bxi \in \mathcal{C}$ \citep[see][for further discussion]{LopezLopera2017FiniteGPlinear}.
	
	The effect of different constraints on samples from the prior $\Uplambda_m(\cdot)$ can be seen in Figure \ref{fig:lineqGPexamples}. Here we set $m = 100$ and use a squared-exponential (SE) covariance function\footnote{SE covariance function: $k(t,t') = \sigma^2 \exp(-\frac{(t-t')^2}{2 {\ell}^2})$.} with covariance parameters $\sigma^2 = 1$, $\ell = 0.2$. The samples were generated via Hamiltonian Monte Carlo (HMC) \citep{Pakman2014Hamiltonian}.
	
	\subsection{Application to 1D GP-Modulated Cox Processes}
	\label{subsec:finiteGaussianCoxProcess}	
	
	The key challenge in building GP-modulated Cox processes is the evaluation of the integral in the intensity measure. 
	By considering $\Uplambda_m(\cdot)$ as the intensity of the Cox process, the intensity measure \eqref{eq:intensitymeasure} becomes
	\begin{align*}
	\mu_m
	= \int_{0}^1 \Uplambda_m(x) \,\dee x 
	= \int_{0}^1 \sum_{j=1}^m \phi_j(x) \xi_j \,\dee x
	= \sum_{j=1}^m c_j \xi_j,
	\end{align*}
	where $c_1 = c_m = \frac{\Delta_m}{2}$ and $c_j = \Delta_m$ for $1 < j < m$.
	The likelihood of $(N=n, X_1 = x_1, \cdots, X_n = x_n)$ is
	\begin{align}
	f_{(N, X_1, \cdots, X_n)|\{\xi_1, \cdots, \xi_m\}}(n, x_1, \cdots, x_n)
	= \frac{1}{n!} \exp\bigg(- \sum_{j=1}^m c_j \xi_j \bigg) \prod_{i=1}^{n} \sum_{j=1}^{m} \phi_j (x_i) \xi_j.
	\label{eq:uncondJointFiniteCoxLikelihoodXi2}
	\end{align}
	Since \eqref{eq:uncondJointFiniteCoxLikelihoodXi2} depends on r.v.'s $\xi_1, \cdots, \xi_m$, it can be approximated using samples of $\Bxi$. To estimate the covariance parameters $\Btheta$ of the vector $\Bxi$, we can use stochastic global optimisation \citep{Jones1998EGO}.
	
	\subsection{Extension to Higher Dimensions}
	\label{subsec:finiteGaussianProcess_HD}
	The approximation in \eqref{eq:finApprox} can be extended to grids in $d$ dimensions by tensorisation. For ease of notation, we assume the same number of knots $m$ and knot-spacing $\Delta_m$ in each dimension, but the generalisation to different $m_1, \cdots, m_d$ or $\Delta_{m_1},\cdots,\Delta_{m_d}$ is straightforward. Consider $\Bx = (x_1, \cdots, x_d) \in [0, 1]^d$, and a set of knots per dimension $(t_1^1, \cdots, t_{m}^1), \cdots, (t_1^d, \cdots, t_{m}^d)$. Then $\Uplambda_m$ is given by
	\begin{align}
	\Uplambda_m (\Bx) = \sum_{j_1, \cdots, j_d = 1}^{m} \Bigg[\prod_{i = \{1, \cdots, d\}} \phi_{j_i}^i (x_i)\Bigg] \xi_{j_1, \cdots, j_d},
	\label{eq:constrFinApprox} 
	\end{align}
	where $\xi_{j_1, \cdots, j_d} := \Uplambda(t_{j_1}, \cdots, t_{j_d})$ and $\phi_{j_i}^i$ are the hat basis functions defined in \eqref{eq:hatbasisfun}. Inequality constraints can be imposed as in \citet{LopezLopera2017FiniteGPlinear}.
	By substituting \eqref{eq:constrFinApprox} in \eqref{eq:intensitymeasure}, we obtain
	\begin{equation*}
	\mu_{m}
	= \int_{0}^1 \Uplambda_{m} (\Bx) \,\dee \Bx
	= \sum_{j_1, \cdots, j_d = 1}^{m} \Bigg[\prod_{i = \{1, \cdots, d\}} c_{j_i} \Bigg] \xi_{j_1, \cdots, j_d},
	\end{equation*}
	with $c_{j_i}$ defined as in 1D, and the likelihood is
	\begin{align}
	f_{(N, X_1, \cdots, X_n)|\Bxi}(n, \Bx_1, \cdots, \Bx_n)
	= \frac{1}{n!}& \exp\Bigg(- \sum_{j_1, \cdots, j_d = 1}^{m} \Bigg[\prod_{i = \{1, \cdots, d\}} c_{j_i} \Bigg] \xi_{j_1, \cdots, j_d} \Bigg) \label{eq:uncondJointFiniteSpatiaCoxLikelihoodXi}\\
	& \times \prod_{i=1}^{n} \sum_{j_1, \cdots, j_d = 1}^{m} \Bigg[\prod_{k = \{1, \cdots, d\}} \phi_{j_i} (x_{i,k})\Bigg] \xi_{j_1, \cdots, j_d}, 	\nonumber
	\end{align}
	where $x_{i,k}$ is the $k$-th component of the point $\Bx_i$.
	
	Due to the tensor structure of the finite representation, it becomes costly as the dimension $d$ increases. The HMC sampler for truncated multivariate Gaussians from \citet{Pakman2014Hamiltonian} follows the same dynamics as a classical HMC sampler, but the particle ``bounces'' on the boundaries if its trajectory reaches one of the inequality constraints.
	The computational complexity of each iteration scales linearly with the number of inequality conditions (e.g.\ $m^d$ for positiveness constraints) if the iteration does not require any reflection, but also increases with each bounce. Hence, in the best case, the computational complexity is $\mathcal{O}(m^d)$. However, this drawback could be mitigated by using sparse representations of the constraints \citep{Pakman2014Hamiltonian}, or using other types of designs of the knots (e.g.\ sparse designs).

	\section{COX PROCESS INFERENCE}
	\label{sec:CPinference}
	
	Having introduced the model, we now establish an inference procedure for $\Uplambda(\cdot)$ using the approximation $\Uplambda_m(\cdot)$. For readability, we only assume non-negativeness constraints, i.e.\ $\Bxi \geq \Bzero$, but the extension to other types of constraints can be made by constructing a set of linear inequalities of the form $\Bl \leq \BA \Bxi \leq \Bu$, where $\BA$ is a full-rank matrix encoding the linear operations, and $\Bl$ and $\Bu$ are the lower and upper bounds. In that case, results for $\BA \Bxi | \{\Bl \leq \BA \Bxi \leq  \Bu\}$ are similar as for $\Bxi | \{\Bzero \leq \Bxi < \boldsymbol{\infty}\}$, and samples of $\Bxi$ can be recovered from samples of $ \BA \Bxi$, by solving a linear system.
	
	Consider the non-negative Gaussian vector $\Bxi$ and its sample $\Bchi$. The posterior distribution of $\Bxi$ conditioned on a point pattern $(N=n, X_1=x_1, \cdots, X_n = x_n)$ is
	\begin{align}
	f {\Bxi|\{N=n, X_1 = x_1, \cdots, X_n = x_n\}}(\Bchi)
	\propto f_{(N, X_1, \cdots, X_n)|\{\Bxi = \Bchi\}}(n, x_1, \cdots, x_n) \; f_\Bxi(\Bchi),
	\label{eq:CoxPosterior}
	\end{align}
	where the likelihood is defined in \eqref{eq:uncondJointFiniteCoxLikelihoodXi2} and $f_\Bxi(\Bchi)$ is the (truncated) Gaussian density given by
	\begin{equation}
	f_\Bxi(\Bchi) = \frac{\exp\left\{ -\frac{1}{2} \Bchi^\top \BGamma^{-1} \Bchi \right\}}{\int_{0}^{\infty} \exp\left\{ -\frac{1}{2} \Bs^\top \BGamma^{-1} \Bs \right\} \dee \Bs}, \; \mbox{ for } \; \Bchi \geq \Bzero.
	\label{eq:priorXi}
	\end{equation}
	Since the posterior distribution \eqref{eq:CoxPosterior} can be approximated using samples of $\Bxi$, it is possible to infer $\Uplambda_m(\cdot)$ via Metropolis-Hastings.
	
	\subsection{Metropolis-Hastings Algorithm with Truncated Gaussian Proposals}
	\label{subsec:MHTrProposal}
	The implementation of the Metropolis-Hastings algorithm requires a proposal distribution $q$ for the next step in the Markov chain. In practice, Gaussian proposals are often used, leading to the famous {random-walk Metropolis algorithm} \citep{Murphy2012ML}. However, since inequality constraints are not necessarily satisfied using (non-truncated) Gaussian proposals, the standard random walk can suffer from small acceptance rates due to constraint violations. We propose as an alternative a constrained version of the random-walk Metropolis algorithm where inequality conditions are ensured when sampling from the proposal $q$.
	As $\Bxi$ is (non-negative) truncated Gaussian-distributed (with covariance matrix $\BGamma$), we suggest the truncated Gaussian proposal $q$ given by
	\begin{align}
	q(\Bchi^{k+1}|\Bchi^{k})
	= \frac{\exp\left\{ -\frac{1}{2} [\Bchi^{k+1}-\Bchi^{k}]^\top \BSigma^{-1} [\Bchi^{k+1}-\Bchi^{k}] \right\}}{\int_{0}^{\infty} \exp\left\{ -\frac{1}{2} [\Bs-\Bchi^{k}]^\top \BSigma^{-1} [\Bs-\Bchi^{k}] \right\} \dee \Bs},
	\label{eq:truncatedQ}
	\end{align}
	where $\Bchi^{k+1}, \Bchi^{k} \geq \Bzero$ are samples of $\Bxi$ and $\BSigma$ is the covariance matrix. Sampling from $q$ can then be performed via MCMC \citep{Pakman2014Hamiltonian}.
	We use $\BSigma = \eta \BGamma$, where $\eta$ is a scale factor. This has the benefit that we are sampling from a distribution with similar structure to the true one, while $\eta$ controls the step size of the Metropolis-Hastings procedure and can be manually tuned to obtain a trade-off between mixing speed and acceptance rate of the algorithm. The acceptance probability is given by 
	\begin{equation}
	\alpha_k
	= \frac{\widetilde{f}_{\Bxi|\{N=n, X_1 = x_1, \cdots, X_n = x_n\}}(\Bchi^{k+1})}{\widetilde{f}_{\Bxi|\{N=n, X_1 = x_1, \cdots, X_n = x_n\}}(\Bchi^{k})} \times \beta_k ,
	\label{eq:acceptProbXi}
	\end{equation}
	where $\beta_k = 
	q(\Bchi^{k}|\Bchi^{k+1})/q(\Bchi^{k+1}|\Bchi^{k})
	$, and
	\begin{align}
	\widetilde{f}_{\Bxi|\{N=n, X_1 = x_1, \cdots, X_n = x_n\}}(\Bchi)
	= \exp\!\Big(\! -\frac{1}{2} \Bchi^\top \BGamma^{-1} \Bchi - \Bc^\top \Bchi \Big) \prod_{i=1}^{n} \Bphi^\top(x_i) \Bchi
	\label{eq:CoxPosteriorApprox}
	\end{align}
	is the (unnormalised) posterior distribution. $\Bphi(\cdot) = [\phi_1(\cdot), \cdots, \phi_m(\cdot)]^\top$ and $\Bc = [c_1, \cdots, c_m]^\top$ are defined in \eqref{eq:hatbasisfun} and \eqref{eq:uncondJointFiniteCoxLikelihoodXi2}. We now focus on the term $\beta_k$. Since the truncated Gaussian density has the same functional form as the non-truncated one, apart from the differing support and normalising constants, this yields
	\begin{align}
	\hspace{-8pt}
	\beta_k = \frac{ \int_{0}^{\infty} \exp\left\{ -\frac{1}{2} [\Bs-\Bchi^{k}]^\top \BSigma^{-1} [\Bs-\Bchi^{k}] \right\} \dee \Bs}{\int_{0}^{\infty} \exp\left\{ -\frac{1}{2} [\Bs-\Bchi^{k+1}]^\top \BSigma^{-1} [\Bs-\Bchi^{k+1}] \right\} \dee \Bs}.
	\label{eq:propRatio}
	\end{align}
	The orthants $\int_{0}^{\infty} \exp\left\{ -\frac{1}{2} [\Bx-\Bmu]^\top \BSigma^{-1} [\Bx-\Bmu] \right\} \dee \Bx$ cannot be computed in closed form, but they can be estimated via MC \citep{Genz1992numericalcomputation,Botev2017MinimaxTilting}. Algorithm \ref{alg:MHCoxProcess} summarises the implementation of the Metropolis-Hastings algorithm for the Cox process inference using the finite approximation of Section \ref{sec:finiteCoxProcess}.
	
	\begin{algorithm}[t!]
		\caption{Metropolis-Hastings algorithm for Cox process inference with truncated Gaussian proposals}\label{alg:MHCoxProcess}
		\begin{algorithmic}[1]
			\BState Input: $\Bchi^{(0)} \in (\realset{m})^+$, $\BGamma$ (covariance matrix of $\Bxi$), $\eta$ (scale factor).
			\For {$k = 0, 1, 2, \cdots$}
			\State Sample $\Bchi' \sim \normrnd{\Bchi^{(k)}}{\eta \BGamma}$ such that $\Bchi' \in \mathcal{C}_+$.
			\State Compute $\alpha_k$ as in \eqref{eq:acceptProbXi}.
			\State Sample $u_k \sim \mbox{uniform}(0,1)$.
			\State Set new sample to
			\State \hspace{10pt}  $\Bchi^{(k+1)} = \begin{cases} \Bchi', &\mbox{if } \;  \alpha_k \geq u_k \\ \Bchi^{(k)}, &\mbox{if } \; \alpha_k < u_k \end{cases}$.
			\EndFor
			\State Compute $\lambda_m^{(k)}(x) = \sum_{j=1}^m \phi_j(x) \chi_j^{(k)}$ at location $x$ with $\phi_j(\cdot)$ defined in \eqref{eq:hatbasisfun}.
		\end{algorithmic}
	\end{algorithm}
	
	\subsection{Inference with Multiple Observations}
	\label{subsec:multipleObs}
	For $N_o$ independent observations $(X_{\nu,1}, \cdots, X_{\nu,n_\nu})$ with $\nu = 1, \cdots, N_o$, the acceptance probability follows
	\begin{align}
	\hspace{-7pt}
	\alpha_k = \frac{\prod_{\nu=1}^{N_o} {f}_{\Bxi|\{N_\nu=n_\nu, 
			\cdots, X_{\nu,n_\nu} = x_{\nu,n_\nu}\}}(\Bchi^{k+1})}{\prod_{\nu=1}^{N_o} {f}_{\Bxi|\{N_\nu=n_\nu, 
			\cdots, X_{\nu,n_\nu} = x_{\nu,n_\nu}\}}(\Bchi^{k})} \; \beta_k,
	\label{eq:acceptProbXiMulti}
	\end{align}
	with posterior ${f}_{\Bxi|\{N_\nu = n_\nu, X_{\nu,1} = x_{\nu,1}, \cdots, X_{\nu,n_\nu} = x_{\nu,n_\nu}\}}$ and $\beta_k$ given by \eqref{eq:CoxPosterior} and \eqref{eq:propRatio}. Then, Algorithm \ref{alg:MHCoxProcess} can be used with \eqref{eq:acceptProbXiMulti}.

	\begin{figure*}[t!]
		\centering
		\hskip 3ex $N_o = 1$ \hskip 26ex $N_o = 10$ \hskip 25.5ex $N_o = 100$ \\ \vspace{-5pt}
		\subfigure{\includegraphics[width = 0.325\textwidth]{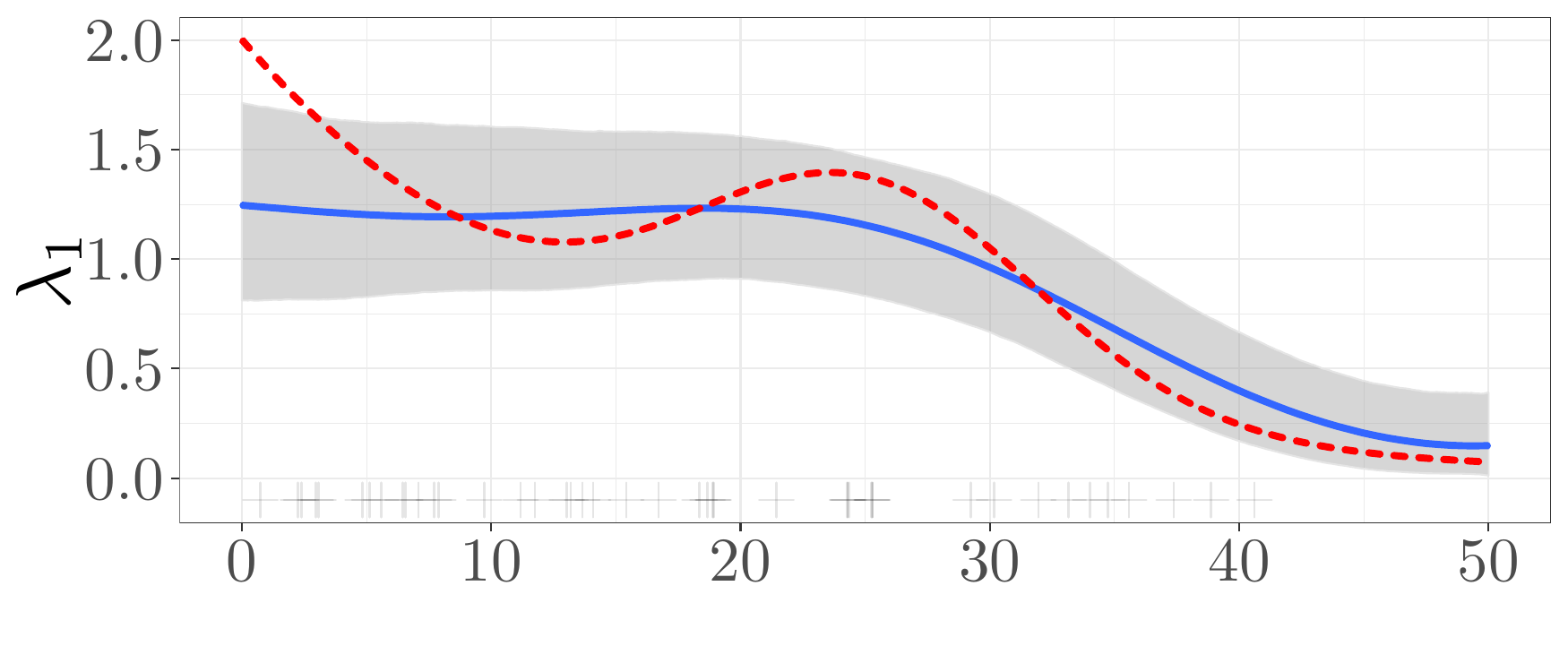}}
		\subfigure{\includegraphics[width = 0.325\textwidth]{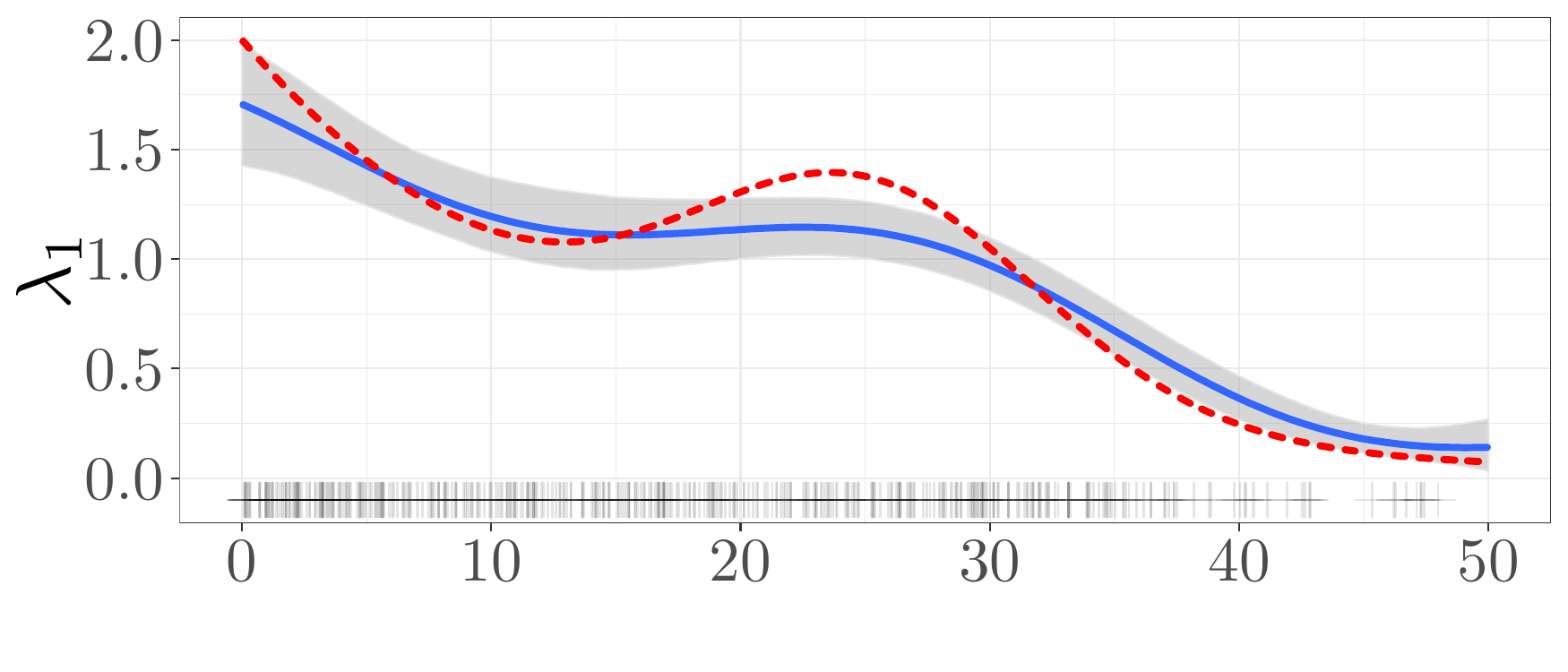}}
		\subfigure{\includegraphics[width = 0.325\textwidth]{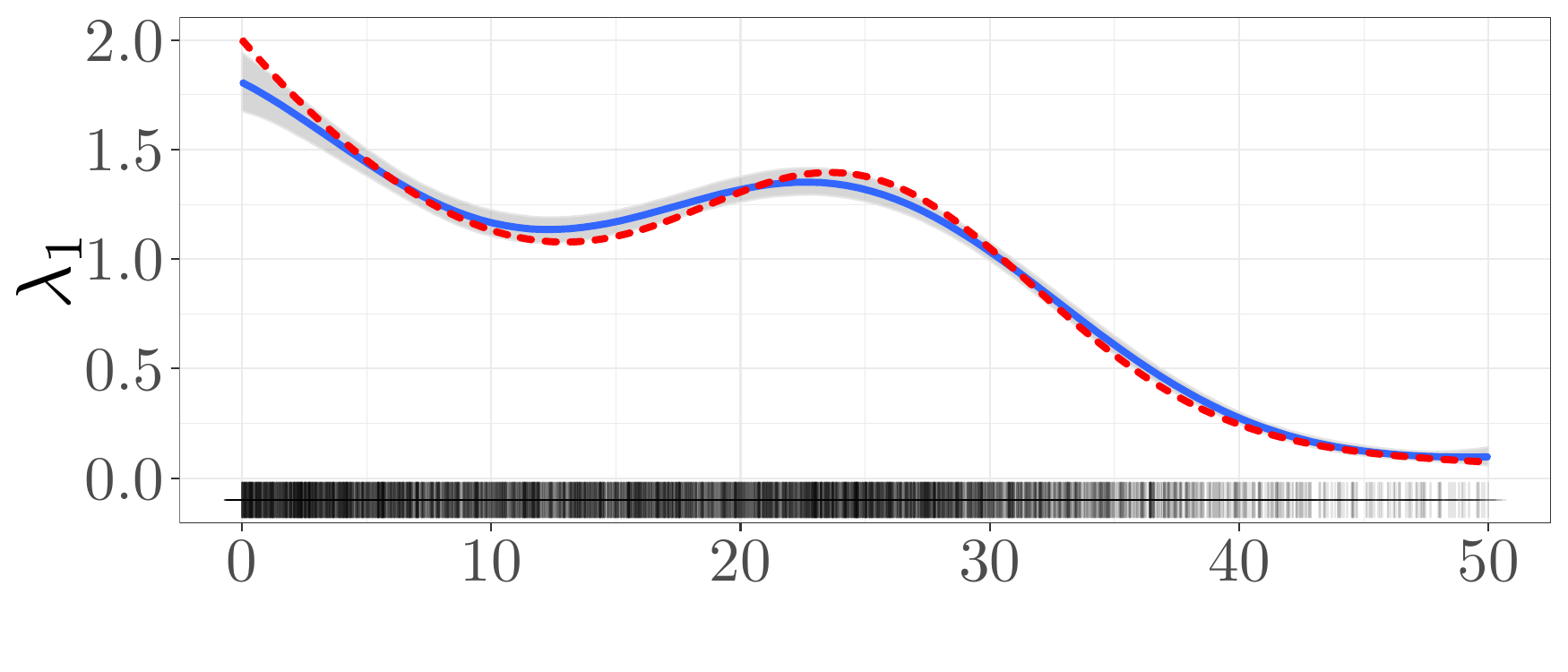}}
		\vspace{-15pt}
		
		\subfigure{\includegraphics[width = 0.325\textwidth]{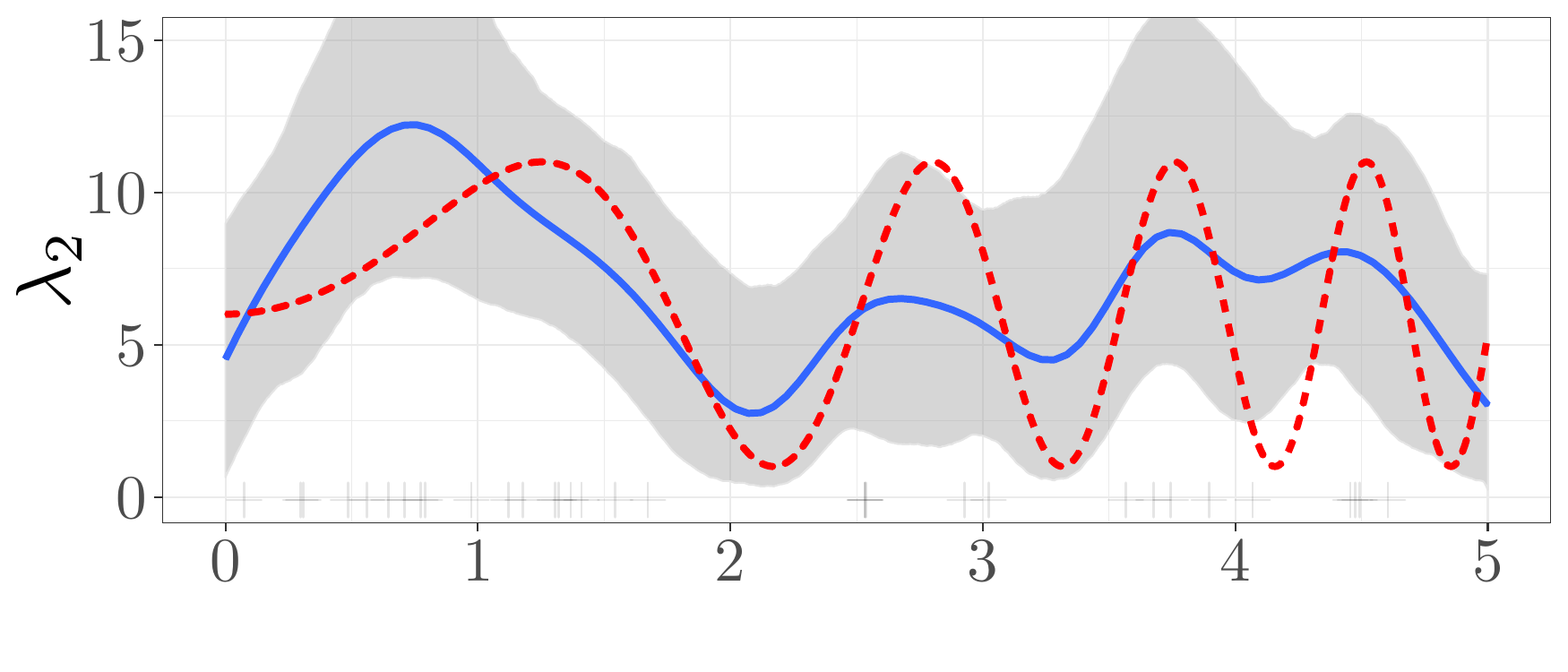}}
		\subfigure{\includegraphics[width = 0.325\textwidth]{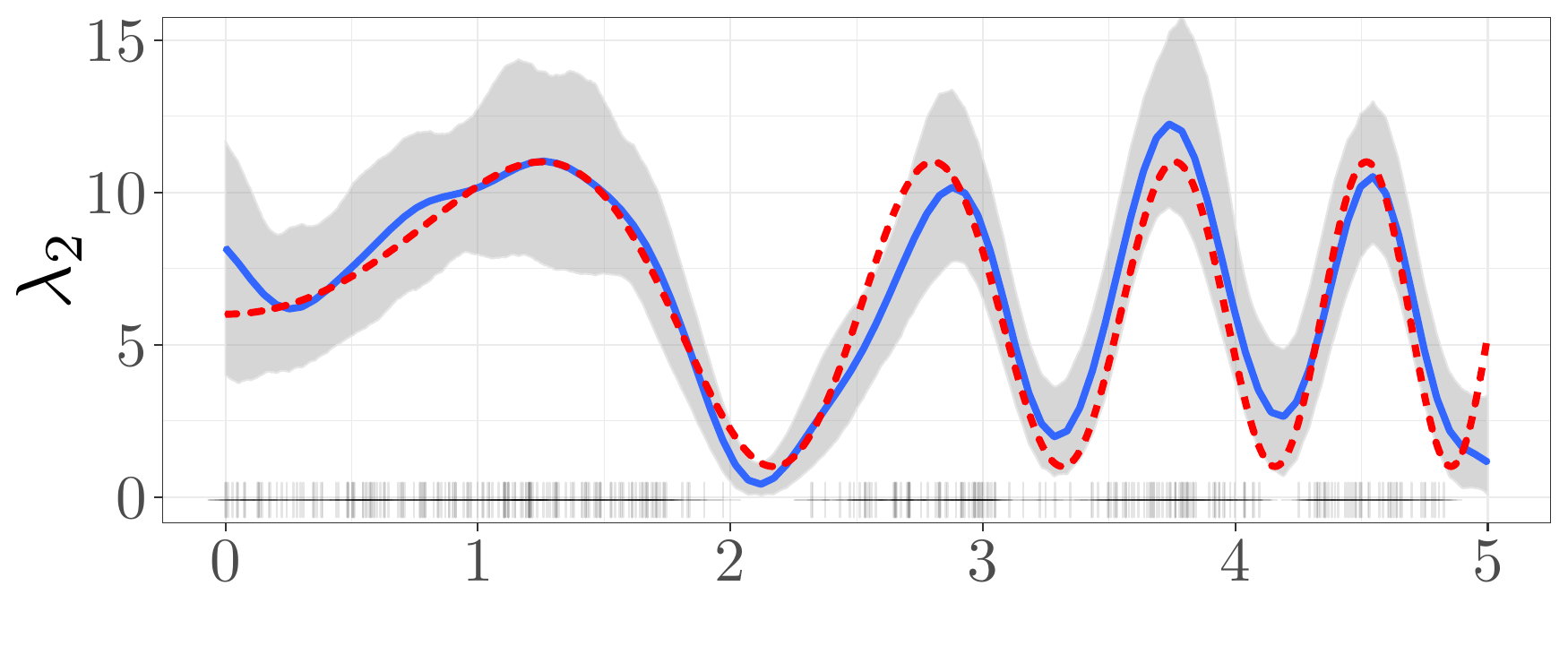}}
		\subfigure{\includegraphics[width = 0.325\textwidth]{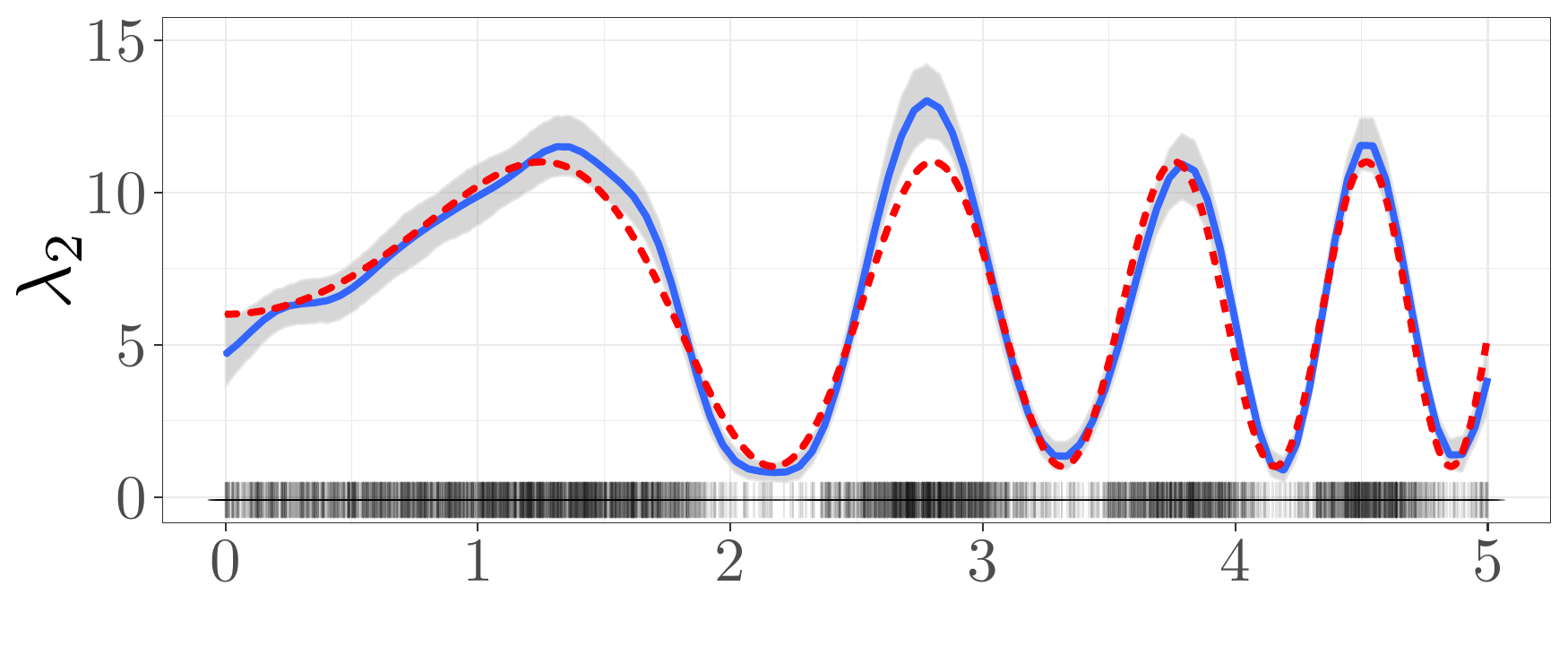}}
		\vspace{-15pt}
		
		\subfigure{\includegraphics[width = 0.325\textwidth]{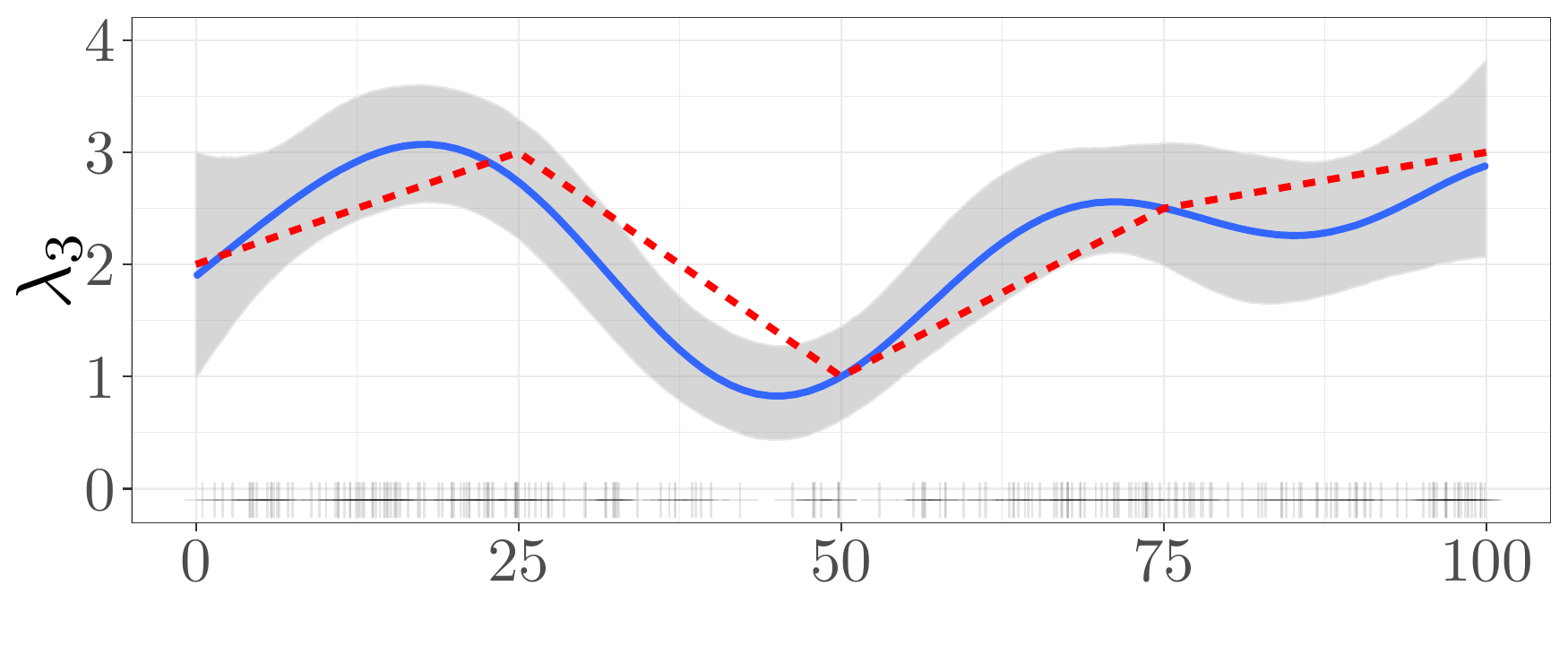}}
		\subfigure{\includegraphics[width = 0.325\textwidth]{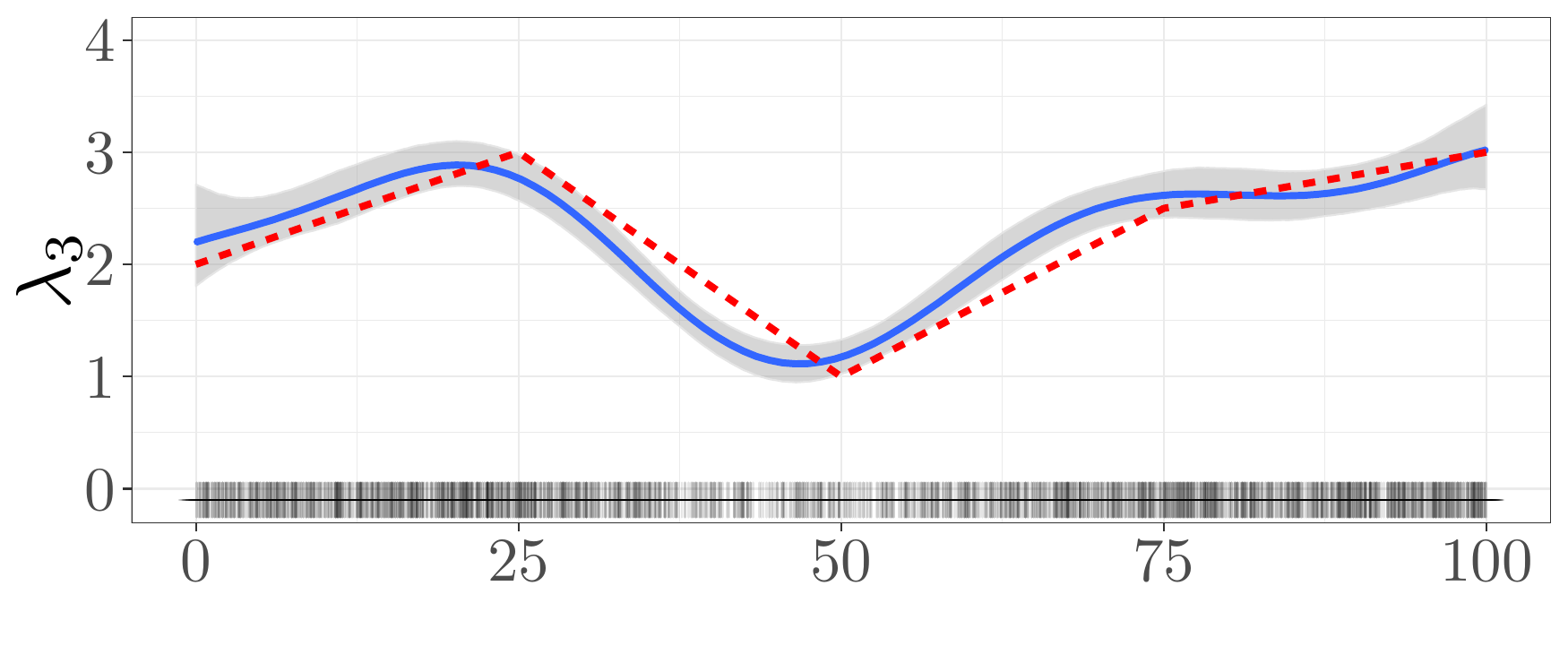}}
		\subfigure{\includegraphics[width = 0.325\textwidth]{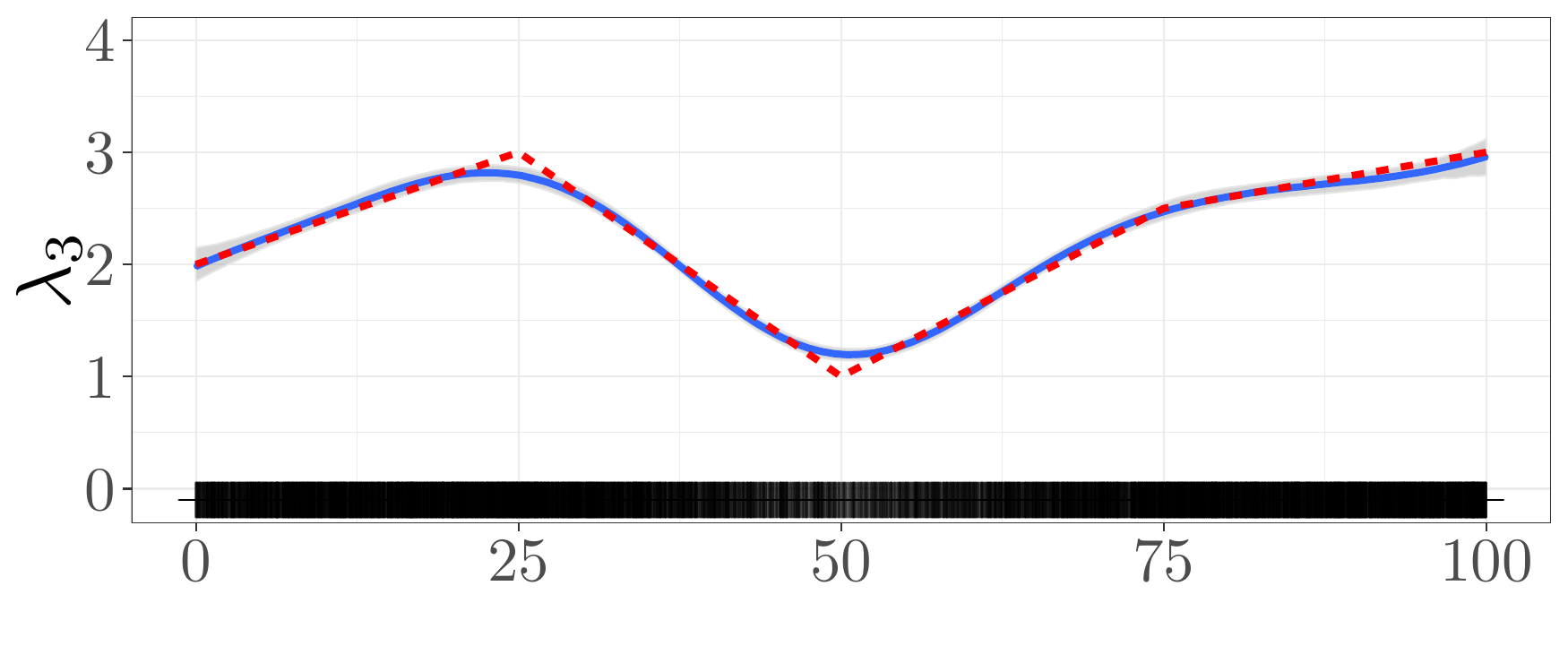}}				
		\caption{Inference results with multiple observations ($N_o = 1, 10, 100$) using the toy examples from \citet{Adams2009CPs}. Each panel shows the point patterns (black crosses), the true intensity $\lambda$ (red dashed lines) and the intensity inferred by the finite approximation of GP-modulated Cox processes (blue solid lines). The estimated 90\% confidence intervals of the finite approximation are shown in grey.}
		\label{fig:toyTiAdamsExamples}
	\end{figure*}	
	\section{EMPIRICAL RESULTS}
	\label{sec:infResults} 
	We test the performance of the finite approximation of GP-modulated Cox process on 1D and 2D applications. In the following, we use the squared-exponential covariance for the Gaussian vector $\Bxi$ so that we can compare to \citet{Lloyd2015CPs}. We estimate the covariance parameters $\Btheta = (\sigma^2, \ell)$ by maximising the likelihood \eqref{eq:uncondJointFiniteCoxLikelihoodXi2}. For all numerical experiments, we fix $m$ such that we obtain accurate resolutions of the finite representations while minimising the cost of MCMC \citep[see][for discussion about the convergence of the finite-dimensional approximation of GPs]{Bay2016KimeldorfWahba,Maatouk2017GPineqconst}.\footnote{We tested our model for various values of $m$, observing that, after a certain value, inference results are unchanged. As a rule of thumb, the number of knots per dimension can be set to $m_i = 10 \cdot \operatorname{range}(\mathcal{S}_i)/\ell_i$ for $i = 1, \cdots, d$.
	}
	For simulating $\Bxi$, we use the exact HMC sampler proposed by \citet{Pakman2014Hamiltonian}. To approximate the Gaussian orthant probabilities from \eqref{eq:propRatio}, we use the estimator proposed by \citet{Botev2017MinimaxTilting} using $200$ MC samples.  We run Algorithm \ref{alg:MHCoxProcess} with a scale factor $\eta$ between $10^{-3}$ and $10^{-4}$ for a good trade-off between the mixing speed and the acceptance rate for each experiment.\footnote{We observed convergence of Algorithm \ref{alg:MHCoxProcess} for a wide range of values of $\eta \in [10^{-5},10^{-2}]$. Fine-tuning of $\eta$ can help the experiment run faster by gradually increasing $\eta$ until the sampling mixes well.} The number of discarded burn-in samples until the Markov chains became stationary varied between $10^3$ and $10^4$ samples. The code was implemented in the R programming language based on the package \texttt{lineqGPR} \citep{LopezLopera2018LineqGPR}.
	
	\subsection{Examples with Multiple Observations}
	\label{subsec:toyExamplesTiAdams}
	Here, we test our approach using the three toy examples proposed by \citet{Adams2009CPs}, 
	\begin{align*}
	\lambda_1(x) &= 2 \exp\{-x/15\} + \exp\{-[(x-25)/10]^2\},\\
	\lambda_2(x) &= 5 \sin(x^2) + 6, \\
	\lambda_3(x) &= \text{piecewise linear through } (0, 2), (25, 3), (50, 1), (75, 2.5) \text{ and } (100, 3).
	\end{align*}
	The domains for $\lambda_1, \lambda_2$ and $\lambda_3$ are $\mathcal{S}_1 = [0, 50]$, $\mathcal{S}_2 = [0, 5]$ and $\mathcal{S}_3 = [0, 100]$, respectively.
	
	Figure \ref{fig:toyTiAdamsExamples} shows the inference results using $N_o = 1, 10, 100$ observations sampled from the ground truth. With increasing number of observations the inferred intensity converges to the ground truth. Here, we fixed $m = 100$ and $\eta = 10^{-3}$.
	
	In Table \ref{tab:Qcomparisson}, we assess the performance of our approach under non-negativeness constraints (cGP-{\tiny$\mathcal{C}_+$}). We compare our inference results to the ones obtained with a log-Gaussian process (log-GP) modulated Cox process \citep{Moller2001LogGaussianCPs} and Variational Bayes for Point Processes (VBPP) \citep{Lloyd2015CPs} using the $Q^2$ criterion. This criterion is defined as $Q^2 = 1 - \operatorname{SMSE}(\lambda(\cdot),\widehat{\lambda}(\cdot))$, where $\operatorname{SMSE}$ is the standardised mean squared error \citep{Rasmussen2005GP}. $Q^2$ is equal to one if the inferred $\widehat{\lambda}(\cdot)$ is exactly equal to the true $\lambda(\cdot)$, zero if it is equal to the average intensity $\overline{\lambda}$, and negative if it performs worse than $\overline{\lambda}$. We compute the $Q^2$ indicator on a regular grid of 1000 locations in $\mathcal{S}$. Then, we compute the mean $\mu$ and one standard deviation $\sigma$ of the $Q^2$ results across $20$ different replicates. Table \ref{tab:Qcomparisson} shows that our approach outperforms its competitors, with consistently higher means of the $Q^2$ results and lesser dispersion $\sigma$.
	
	We assess the computational cost of our approach using the third toy example $\lambda_3$ for $N_o = 100$ (which has the largest number of events with on average $22500$ events in total). Obtaining one sample using our approach takes around $60$ milliseconds, and generating all $10^4$ samples takes $10$ minutes in total (in contrast to the $18$ minutes required by VBPP).\footnote{These experiments were executed on a single core of an Intel\textsuperscript{\textregistered} Core\textsuperscript{TM} i7-6700HQ CPU.} The multivariate effective sample size (ESS) \citep{Flegal2017mcmcse} was estimated at $322$, corresponding to an effective sampling rate of $0.536 \; \mathrm{s}^{-1}$.
	
	\begin{table}[t!]
		\centering
		\caption{$Q^2$ results for the toy examples of Figure \ref{fig:toyTiAdamsExamples}, averaged over 20 ($^\dagger$10) replicates. Our results (cGP-{\tiny$\mathcal{C}_+$}) are compared to results for \citet{Moller2001LogGaussianCPs} (log-GP) and \citet{Lloyd2015CPs} (VBPP).}
		\label{tab:Qcomparisson}
		\begin{tabular}{ccr@{$\pm$}cr@{$\pm$}cr@{$\pm$}c}
			\toprule
			\multirow{2}{*}{Toy} & \multirow{2}{*}{$N_o$} & \multicolumn{6}{c}{$Q^2$ ($\mu \pm \sigma$) [$\%$]} \\
			& &  \multicolumn{2}{c}{log-GP} & \multicolumn{2}{c}{VBPP} & \multicolumn{2}{c}{cGP-{\tiny$\mathcal{C}_+$}} \\
			\midrule
			\multirow{3}{*}{$\lambda_1$}
			& 1 & 51.2 & 30.1 & 51.9 & 26.1 & \textbf{65.7} & \textbf{14.3} \\
			& 10  & 95.1 & 3.9 & 94.6 & 3.7 & \textbf{95.4} & \textbf{2.3} \\
			& 100 & \textbf{99.5} & \textbf{0.2} & \textbf{99.5} & \textbf{0.3} & \textbf{99.5} & \textbf{0.3} \\
			\midrule
			\multirow{3}{*}{$\lambda_2$}
			& 1 & -35.2 & 43.4 & -1.1 & 28.8 & \textbf{0.7 }& \textbf{24.0} \\
			& 10  & 72.6 & 9.1 & 71.7 & 10.4 & \textbf{81.9} & \textbf{7.4} \\
			& 100 & 95.4 & 0.7 & 92.1 & 3.9 & \textbf{97.8} & \textbf{0.6} \\
			\midrule
			\multirow{3}{*}{$\lambda_3$}
			& 1 & 49.2 & 22.6 & 49.5 & 29.9 & \textbf{58.1} & \textbf{21.4} \\
			& 10  & 91.7 & 4.4 & 93.8 & 2.8 & \textbf{94.3} & \textbf{2.5} \\
			& 100 & 98.4 & 0.4 & \textbf{98.9} & \textbf{0.3}$^\dagger$ & \textbf{98.8} & \textbf{0.3} \\
			\bottomrule
		\end{tabular}
	\end{table}
	
	\subsection{Modelling Hazard Rates in Renewal Processes}
	\label{subsec:renewalProcesses}
	
	Poisson processes have been extended to model renewal processes where intensity functions are seen as hazard rates defining the probability that an operating object fails \citep{Serfozo2009RenewalProcesses,Cha2018SpatPointReliability}. However, in many application, e.g.\ reliability engineering and survival analysis, hazard rates exhibit monotonic behaviours describing the degradation of items or lifetime of organisms. For example, the hazard functions for the failure of many mechanistic devices and the mortality of adult humans tend to exhibit monotonic behaviours. Thus, taking monotonicity constraints into account in renewal processes is crucial for the study of many applications. Moreover, it is known that introducing monotonicity information in GPs can lead to more realistic uncertainties \citep{Riihimaki2010GPwithMonotonicity,Maatouk2017GPineqconst}.
	
	As discussed in Section \ref{sec:CoxProcess}, some renewal processes can be seen as Cox processes under certain conditions. In order to demonstrate that we can model other types of point patterns, here we use two toy examples where hazard rates are known to be monotonic. Both examples are inspired by two classical renewal process: Weibull process and Gamma process.
	\begin{figure*}[t!]
		\centering
		\hskip 3ex cGP-{\tiny$\mathcal{C}_+$} \hskip 26ex cGP-{\tiny$\mathcal{C}_+^{\downarrow}$} \hskip 25.5ex cGP-{\tiny$\mathcal{C}_+^{\overset{\downarrow}{\smile}}$} \\ \vspace{-5pt}
		\hskip -2.55ex
		\subfigure{\includegraphics[width = 0.325\textwidth]{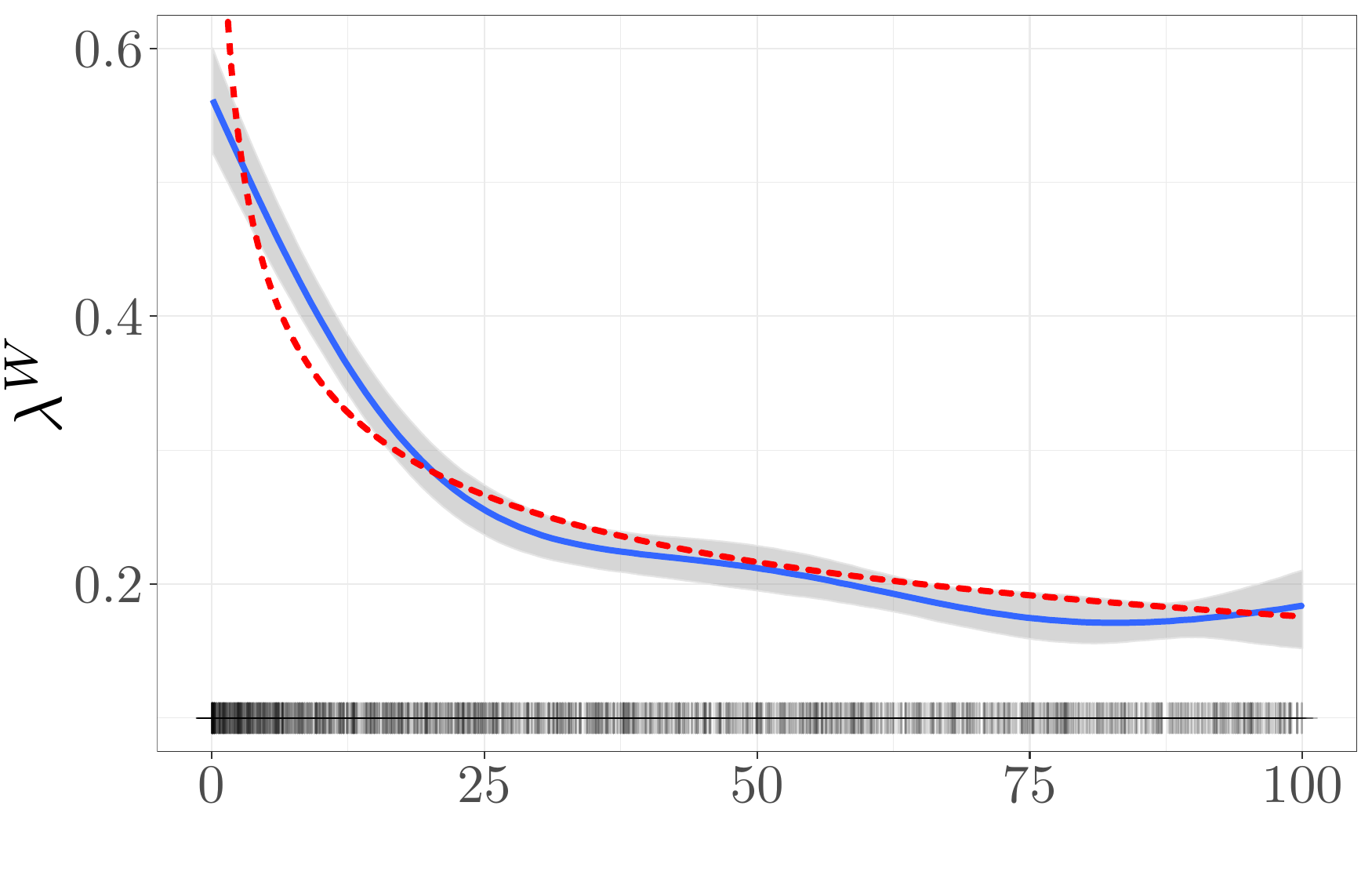}}
		\subfigure{\includegraphics[width = 0.325\textwidth]{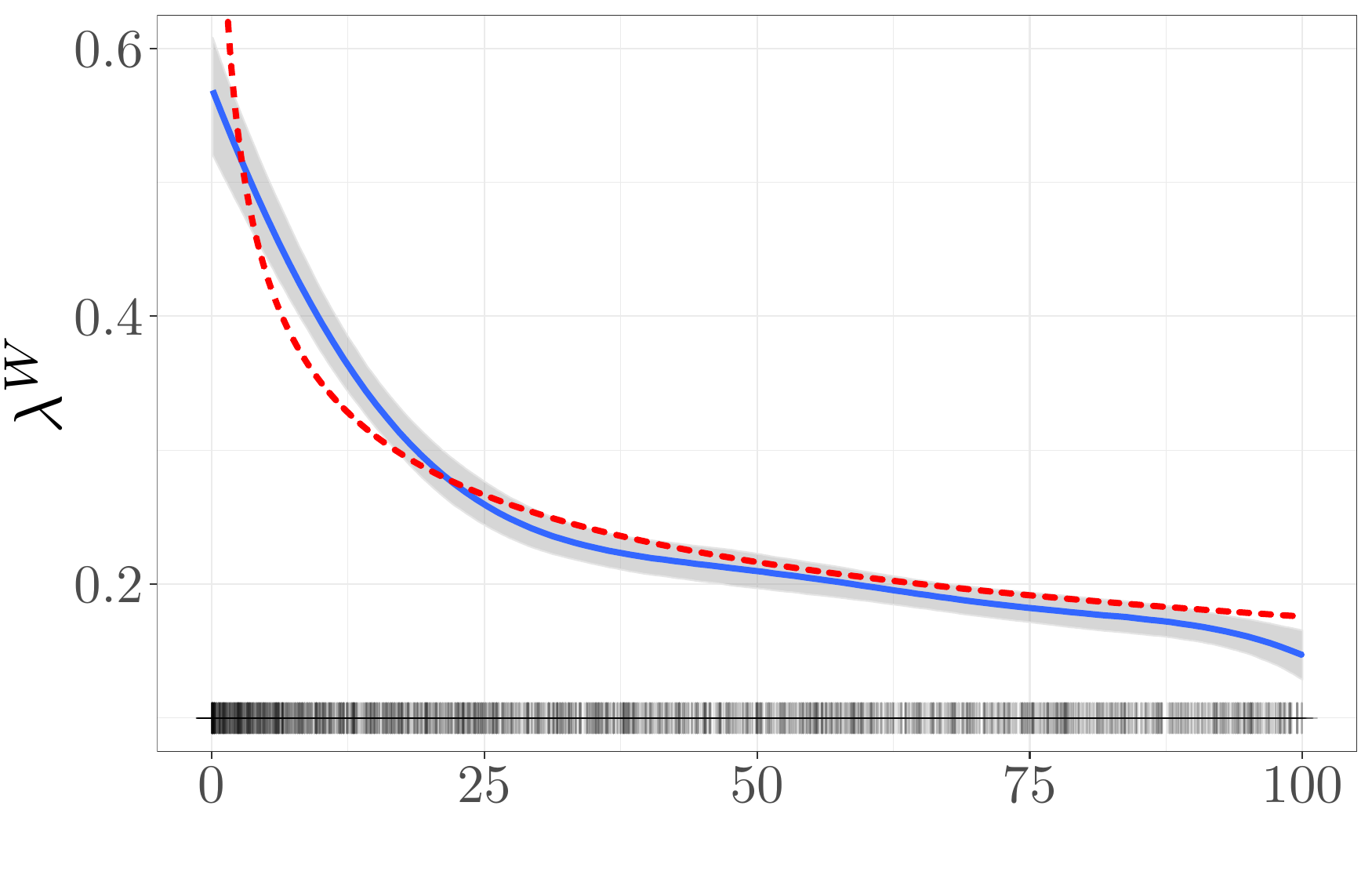}}
		\subfigure{\includegraphics[width = 0.325\textwidth]{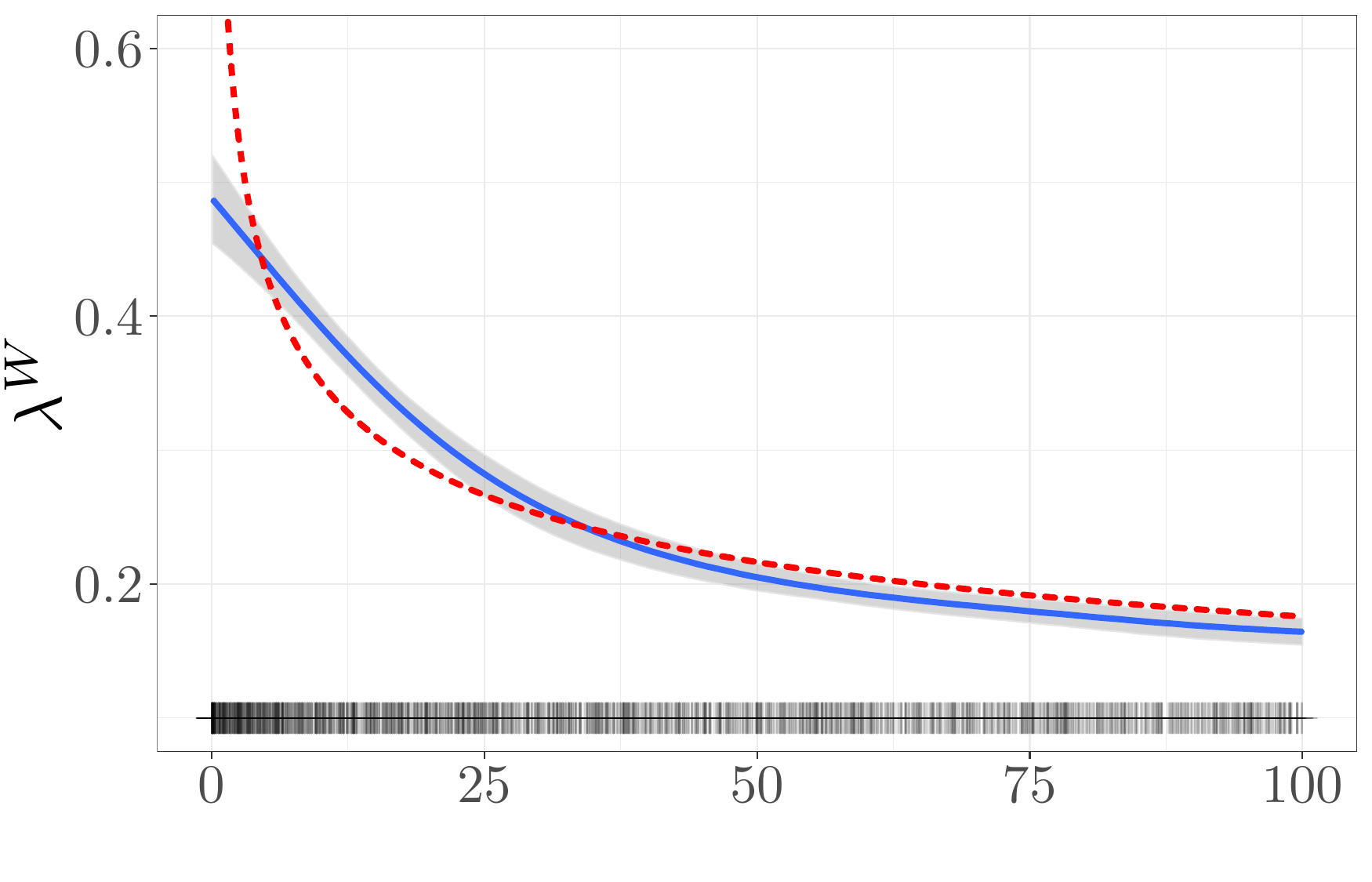}}
		
		\hskip 3ex cGP-{\tiny$\mathcal{C}_+$} \hskip 26ex cGP-{\tiny$\mathcal{C}_+^{\uparrow}$} \hskip 25.5ex cGP-{\tiny$\mathcal{C}_+^{\overset{\uparrow }{\frown}}$}  \\ \vspace{-5pt}
		\subfigure{\includegraphics[width = 0.325\textwidth]{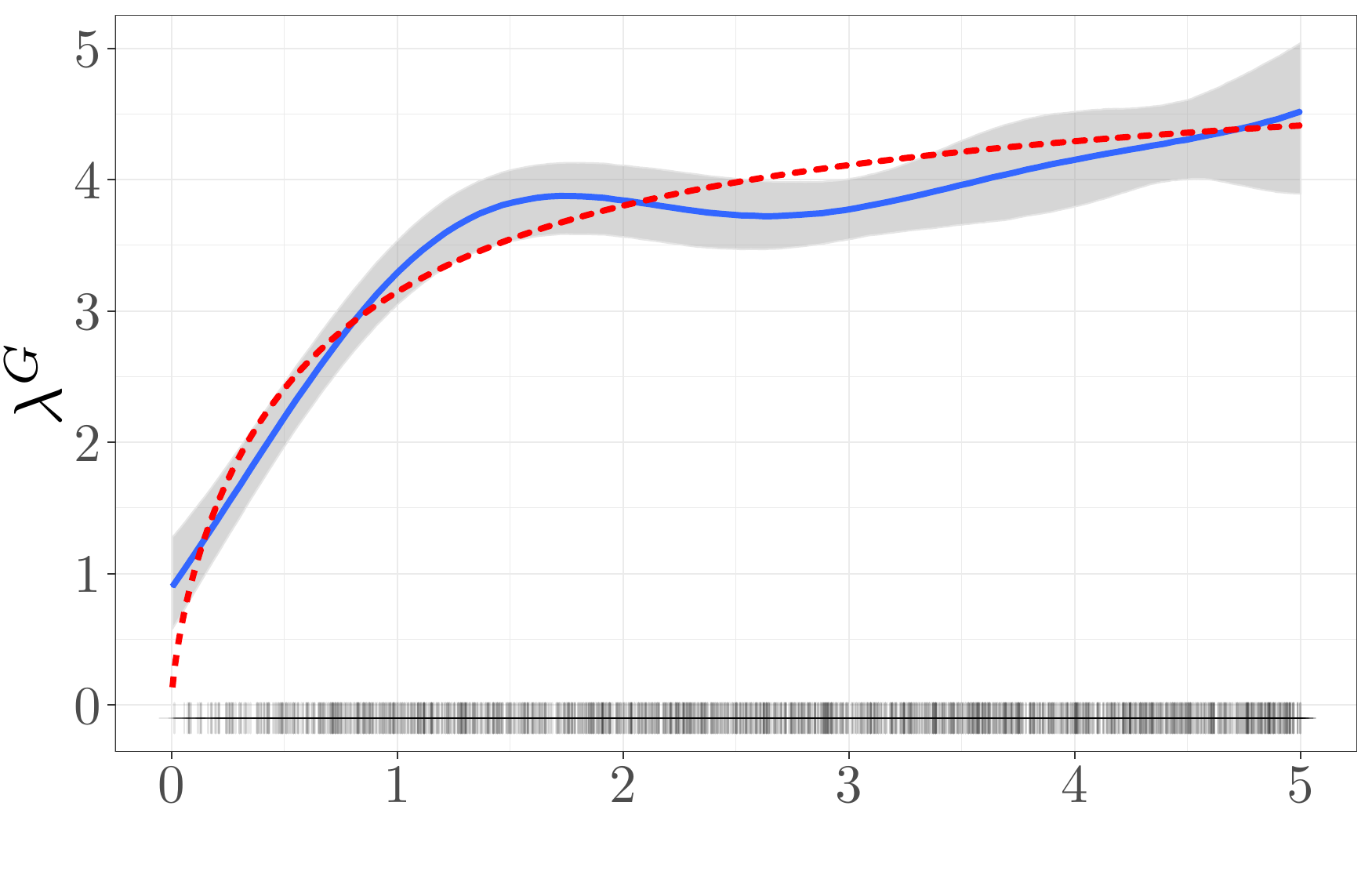}}
		\subfigure{\includegraphics[width = 0.325\textwidth]{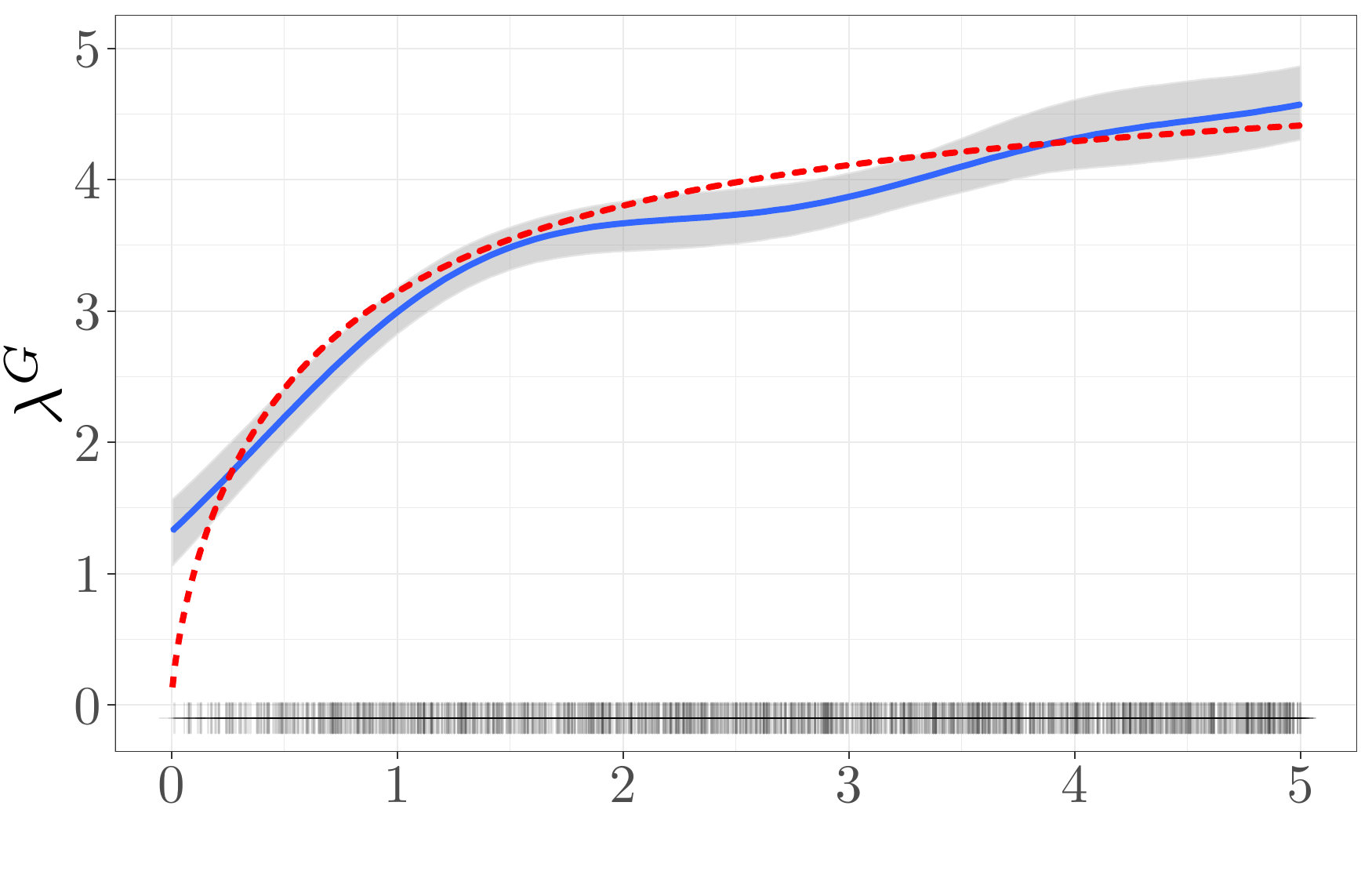}}
		\subfigure{\includegraphics[width = 0.325\textwidth]{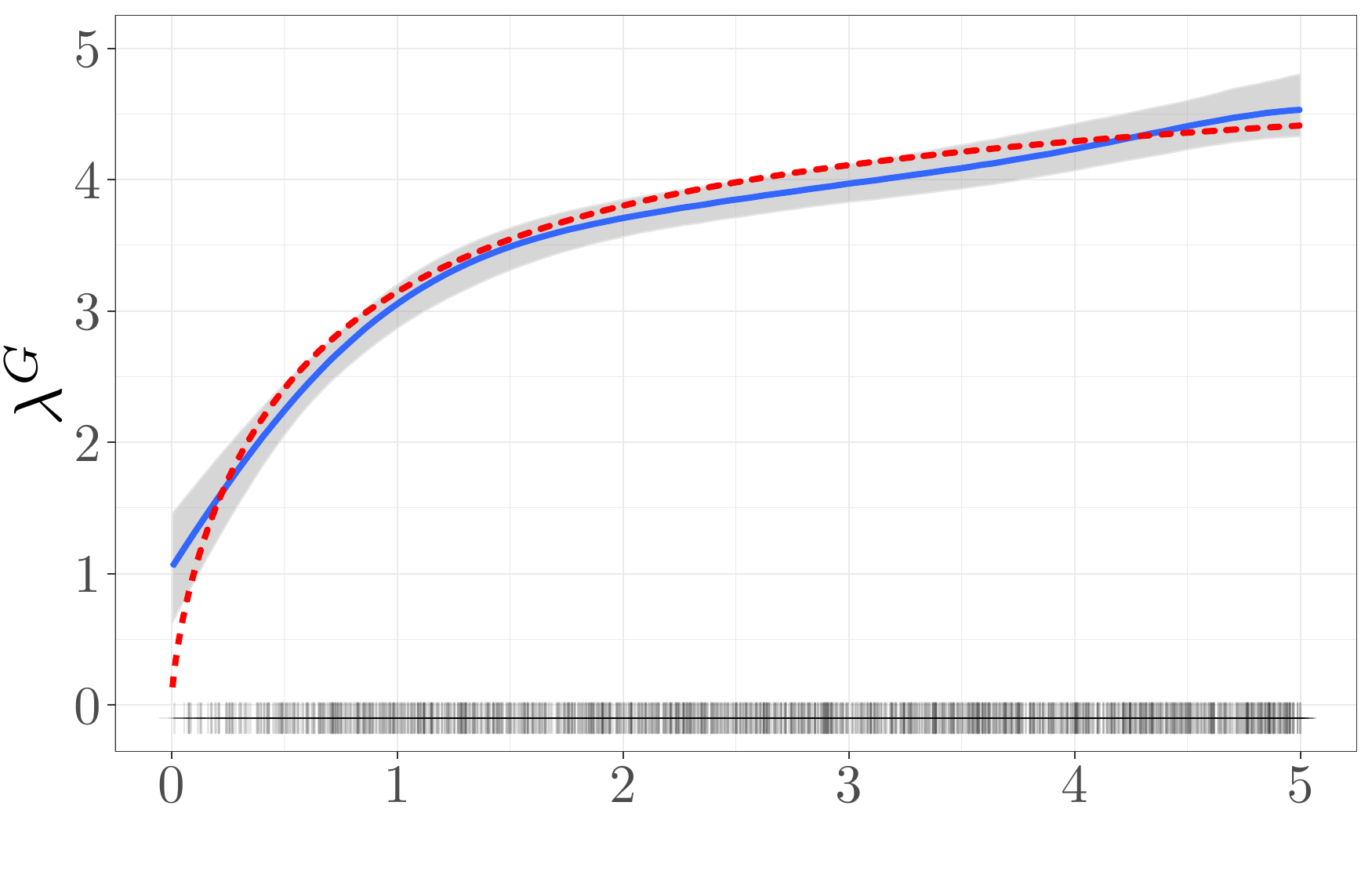}}
		\vspace{-0.5em}
		\caption{Renewal inference examples under different inequality constraints using $N_o = 100$ and $m = 100$. Inference results are shown for (top row) a Weibull renewal process with $\alpha = 1$ and $\beta = 0.7$, and (bottom row) a Gamma renewal process with $\alpha = 5$ and $\beta = 1.7$. The panel description is the same as in Figure \ref{fig:toyTiAdamsExamples}.}
		\label{fig:toyRenewalExamples}
	\end{figure*}
	
	For the first class, the Weibull hazard function is
	\begin{equation}
	\lambda^W(x) = \alpha \beta x^{\beta-1} \; \mbox{ for } \; x\geq 0,
	\label{eq:WeibullIntensity}
	\end{equation}
	where $\alpha$ and $\beta$ are the scale and shape parameters, respectively. Depending on $\beta$, $\lambda^W$ can be either non-increasing ($0 < \beta < 1$), constant ($\beta = 1$), or non-decreasing ($\beta > 1$). Moreover, for $\beta \in (0,1]$, the Weibull renewal process can be seen as a Cox process \citep{Yannaros1988GamaProcesses}. For numerical experiments, we consider the case of non-increasing conditions in the domain $\mathcal{S} = [0, 100]$ by fixing $\alpha = 1$ and $\beta = 0.7$ (see Figure \ref{fig:toyRenewalExamples}). We test our framework using $N_o = 100$ observations from $\lambda^W$, and we consider non-negativeness conditions, with (cGP-{\tiny$\mathcal{C}_+^\downarrow$}) or without (cGP-{\tiny$\mathcal{C}_+$}) taking into account the non-increasing constraint. We also consider the case where $\lambda^W$ is non-increasing and convex (cGP-{\tiny$\mathcal{C}_+^{\overset{\downarrow}{\smile}}$}). 
	
	For the Gamma class, the hazard function is given by
	\begin{equation}
	\lambda^G(x) = \frac{\alpha \; x^{\beta-1} e^{-x}}{\Gamma(\beta) - \Gamma_x(\beta)}, \; \mbox{ for } \; x \geq 0,\label{eq:GammaIntensity}
	\end{equation}	
	where $\Gamma(\cdot)$ and $\Gamma_x(\cdot)$ are the Gamma function and the incomplete Gamma function, respectively \citep{Cha2018SpatPointReliability}, and $\alpha$ and $\beta$ are the scale and shape parameters. As for the Weibull process, different behaviours can be obtained using different values of $\beta$. Since similar profiles are obtained for $\beta \in (0,1]$, here we are interested in the case where $\lambda^G$ exhibits non-decreasing constraints ($\beta > 1$). We fix $\mathcal{S} = [0, 5]$, $\alpha = 5$ and $\beta = 1.7$ obtaining a non-decreasing profile as shown in Figure \ref{fig:toyRenewalExamples}. Here, we consider non-decreasing (cGP-{\tiny$\mathcal{C}_+^{\tiny\uparrow}$}), and non-decreasing and concave (cGP-{\tiny$\mathcal{C}_+^{\tiny \overset{\uparrow}{\frown}}$}) constraints. Since $\lambda^G(x) < \alpha$ for $x \in \mathcal{S}$, we add the constraint $\lambda^G \in [0, \alpha]$.
	
	Figure \ref{fig:toyRenewalExamples} shows the inferred intensities of $\lambda^W$ and $\lambda^G$ under the different conditions previously discussed. In both experiments, we fixed $m = 100$ and $\eta = 10^{-4}$. For the Weibull class $\lambda^W$, the performance of all three models, cGP-{\tiny$\mathcal{C}_+$}, cGP-{\tiny$\mathcal{C}_+^{\downarrow}$} and cGP-{\tiny$\mathcal{C}_+^{\overset{\downarrow}{\smile}}$}, tends to be similar. However, the model without monotonicity constraint exhibits undesired oscillations, whereas the other two approaches provide more realistic decreasing profiles and more accurate inference results for $x > 50$. We can also observe that the three models cannot learn the singularity at $x = 0$. Note that the proposed methodology does not make any assumption on the kernel, and it would be possible to consider a covariance function such as $k(x,y)/(xy)$ in order to improve the model behaviour for small and large values of $x$. For the Gamma hazard function $\lambda^G$, one may clearly observe the benefits of adding the non-decreasing and concave constraints, obtaining absolute improvements between $0.8\%$ and $3.5\%$ of the $Q^2$ indicator. Both examples of Figure \ref{fig:toyRenewalExamples} show that the monotonicity and convexity conditions found in certain point processes can be difficult to learn directly from the data. This suggests that including those constraints in the GP prior is necessary to get accurate models with more realistic uncertainties.
	
	\begin{figure}[t!]
		\centering
		\subfigure[\label{subfig:reedwood1}$\ell_1 = \ell_2 = 0.1$]{\includegraphics[width = 0.325\textwidth]{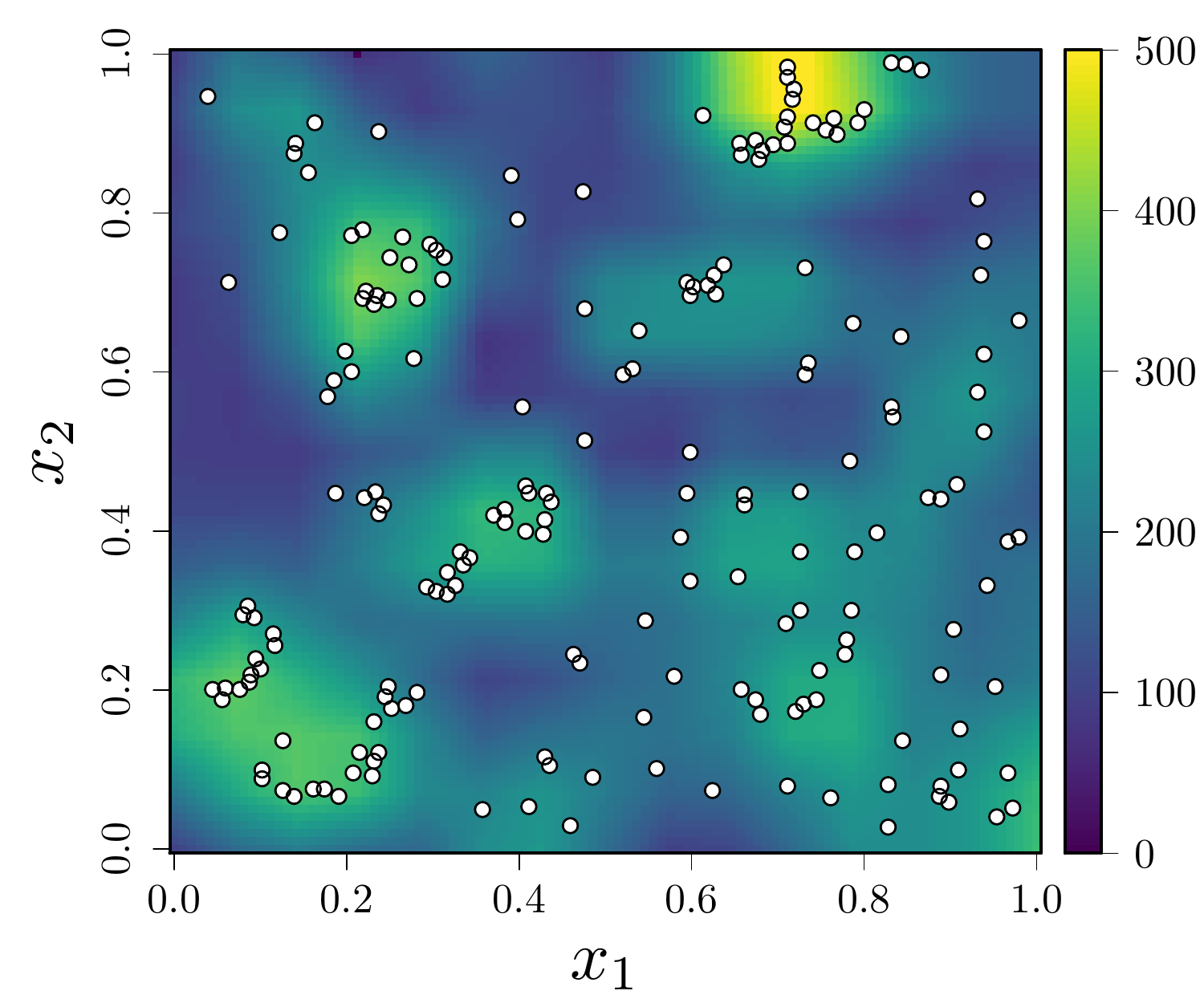}}
		\subfigure[\label{subfig:reedwood2}$\ell_1 = \ell_2 = 0.01$]{\includegraphics[width = 0.325\textwidth]{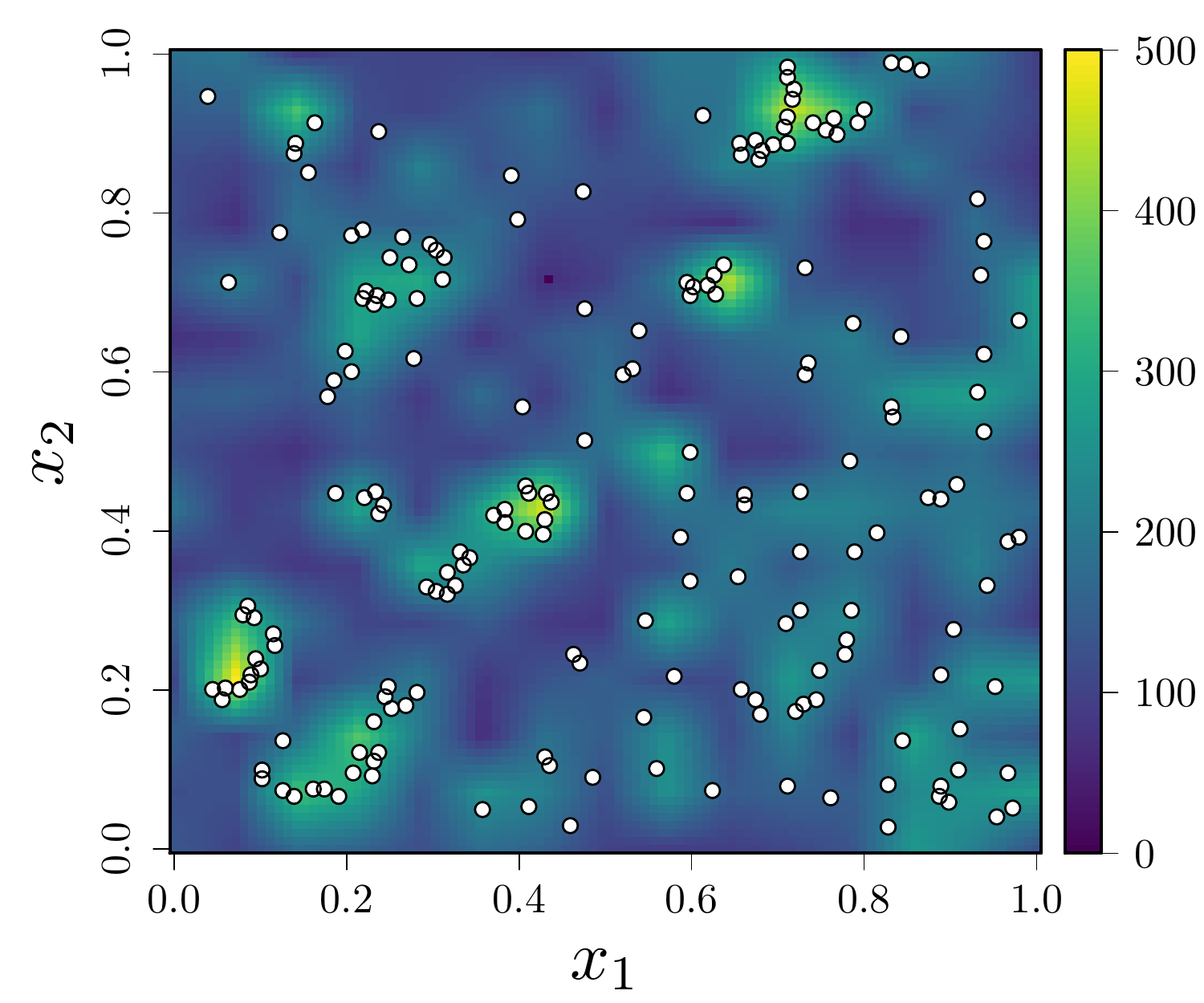}}
		\subfigure[\label{subfig:reedwood3}$\widehat{\ell}_1 = 0.055$, $\widehat{\ell}_2 = 0.084$]{\includegraphics[width = 0.325\textwidth]{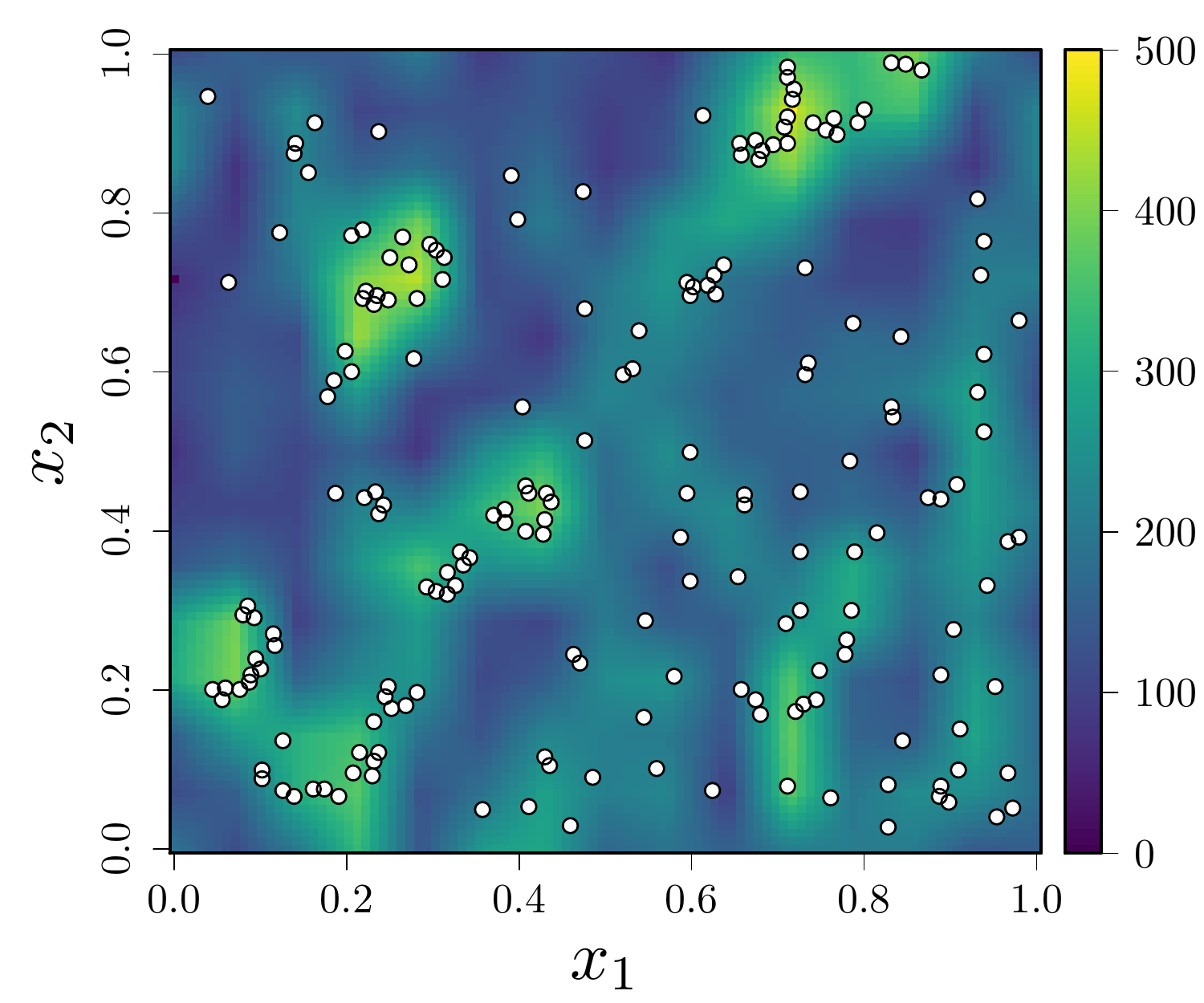}}
		\caption{Inference results of the redwoods data from \citet{Ripley1977SpatialPatterns,Baddeley2015SpatStatsR}. Each panel shows the point pattern (white dots) and the estimated intensity $\lambda(\cdot)$.}
		\label{fig:dataRipley}
		\vspace{-2ex}
	\end{figure}	
	\subsection{2D Redwoods Data}
	\label{subsec:RedwoodsData}
	We now assess the performance of the proposed approach for a 2D spatial problem. We use the dataset provided by \citet{Ripley1977SpatialPatterns} which describes the locations of redwood trees. The dataset contains $n = 195$ events scaled to the unit square (see Figure \ref{fig:dataRipley}). Here we choose $m = 15$, obtaining 225 knots in total, to obtain a good trade-off between resolution and computational cost. We use the product of two SE kernels with covariance parameters $\Btheta = (\sigma^2,{\ell}_1,{\ell}_2)$ as the covariance function of the Gaussian vector $\Bxi$, and we choose $\eta = 10^{-4}$ in Algorithm \ref{alg:MHCoxProcess}. Following the burn-in step, we keep $10^5$ samples for the inference of $\lambda(\cdot)$, yielding a total running time of $7.6$ hours (i.e.\ a sampling rate of approximately $4\; \mathrm{s}^{-1}$).
	
	Figure \ref{fig:dataRipley} shows the normalised inference results for the redwood dataset for different values of the lengthscale parameters. Since in our approach we directly impose the inequality conditions on the Gaussian vector $\Bxi$ instead of using a link function, the interpretation of the lengthscale parameters $({\ell}_1,{\ell}_2)$ are the same as for standard GPs: one can find a trade-off between fidelity and regularity by tuning $\ell$. One can note, from Figures \ref{subfig:reedwood1} and \ref{subfig:reedwood2}, that both profiles tend to properly learn the point patterns but more regularity is exhibited when $\ell_1 = \ell_2 = 10^{-1}$. For the case $\ell_1 = \ell_2 = 10^{-2}$, although the model follows the point patterns, one may observe noisy behaviour in regions without points, e.g.\ around $(x_1, x_2) = (0.30, 0.85)$, as small values of $\ell$ lead to more oscillatory Gaussian random fields.
	Finally, we infer $\lambda(\cdot)$ when the covariance parameters $\Btheta$ are estimated via maximum likelihood using \eqref{eq:uncondJointFiniteSpatiaCoxLikelihoodXi}. According to the estimated lengthscales $({\widehat{\ell}}_1 = 0.055,\  {\widehat{\ell}}_2 = 0.084)$, one can conclude that the estimated intensity $\lambda(\cdot)$ is smoother along the second dimension $x_2$. This is in agreement with the inference results by \citet{Adams2009CPs}, where more variations of $\lambda(\cdot)$ were exhibited across $x_1$.
	
	\section{CONCLUSIONS}
	\label{sec:conclusions}
	
	The proposed model for GP-modulated Cox processes is based on a finite-dimensional approximation of a GP that is constrained to be positive. This approach shows several advantages. First of all, it is based on general linear inequality constraints so it allows us to incorporate more information, such as monotonicity and convexity, in the prior. As seen in the experiments, this appears to be particularly helpful when few data are available. Second, imposing directly the positivity constraint on the GP makes the use of a link function unnecessary. Both the likelihood and the intensity measure can be computed analytically, which is not always the case when using a link function. Finally, the fact that our model is based on a finite-dimensional representation ensures that the computational burden grows linearly with the number of observations.
	
	There are two key elements that make the method work: (a) the finite-dimensional representation of the GP that ensures that the constraints are satisfied everywhere, and (b) the dedicated MCMC proposal distribution based on a truncated normal distribution which allows us to have high acceptance rates compared to a naive multivariate Gaussian proposal.
	
	The main limitation regarding the scaling of the proposed method lies in the dimension of the input space. This is due to the construction by tensorisation of the basis functions used to obtain the finite-dimensional representation. Moreover, our model is also sensitive to three parameters: the dimensionality of the space in which we perform HMC, the number of constraints, and the number of times the HMC particles violate a constraint. However, we believe that these limitations are not inherent to the proposed model and that other types of designs of the knots (e.g.\ sparse designs) could be used in high dimensions.
	
	\subsubsection*{Acknowledgements}	
	{This work was supported by the Chair in Applied Mathematics OQUAIDO (\url{oquaido.emse.fr}, France) and by PROWLER.io (\url{www.prowler.io}, UK). We thank O. Roustant (EMSE) and D. Rulli\`ere (ISFA) for their advice throughout this work.}
	
\bibliographystyle{apa}  
\bibliography{aistats2018arXiv}  

\begin{thebibliography}{}

\bibitem[\protect\astroncite{Adams et~al.}{2009}]{Adams2009CPs}
Adams, R.~P., Murray, I., and MacKay, D.~J. (2009).
\newblock Tractable nonparametric {B}ayesian inference in {P}oisson processes
  with {G}aussian process intensities.
\newblock In {\em ICML}, pages 9--16.

\bibitem[\protect\astroncite{Baddeley
  et~al.}{2006}]{Baddeley2006SpatPointProcess}
Baddeley, A., Gregori, P., Mahiques, J., Stoica, R., and Stoyan, D. (2006).
\newblock {\em Case Studies in Spatial Point Process Modeling}.
\newblock Lecture Notes in Statistics. Springer, New York.

\bibitem[\protect\astroncite{Baddeley et~al.}{2015}]{Baddeley2015SpatStatsR}
Baddeley, A., Rubak, E., and Turner, R. (2015).
\newblock {\em Spatial Point Patterns: Methodology and Applications with R}.
\newblock Chapman \& Hall/CRC Interdisciplinary Statistics. CRC Press, Boca
  Raton, FL.

\bibitem[\protect\astroncite{Bay et~al.}{2016}]{Bay2016KimeldorfWahba}
Bay, X., Grammont, L., and Maatouk, H. (2016).
\newblock {Generalization of the Kimeldorf--Wahba correspondence for
  constrained interpolation}.
\newblock {\em {Electronic Journal of Statistics}}, 10(1):1580--1595.

\bibitem[\protect\astroncite{Botev}{2017}]{Botev2017MinimaxTilting}
Botev, Z.~I. (2017).
\newblock The normal law under linear restrictions: Simulation and estimation
  via minimax tilting.
\newblock {\em Journal of the Royal Statistical Society: Series B},
  79(1):125--148.

\bibitem[\protect\astroncite{Cha and
  Finkelstein}{2018}]{Cha2018SpatPointReliability}
Cha, J. and Finkelstein, M. (2018).
\newblock {\em Point Processes for Reliability Analysis: Shocks and Repairable
  Systems}.
\newblock Springer Series in Reliability Engineering. Springer, New York.

\bibitem[\protect\astroncite{Cox}{1955}]{Cox1955}
Cox, D.~R. (1955).
\newblock Some statistical methods connected with series of events.
\newblock {\em Journal of the Royal Statistical Society: Series B},
  17(2):129--164.

\bibitem[\protect\astroncite{Diggle et~al.}{2013}]{Diggle2013CPGeo}
Diggle, P.~J., Moraga, P., Rowlingson, B., and Taylor, B.~M. (2013).
\newblock Spatial and spatio-temporal log-{G}aussian {C}ox processes: Extending
  the geostatistical paradigm.
\newblock {\em Statistical Science}, 28(4):542--563.

\bibitem[\protect\astroncite{Donner and Opper}{2018}]{Donner2018SigmoidalCPs}
Donner, C. and Opper, M. (2018).
\newblock Efficient {B}ayesian inference of sigmoidal {G}aussian {C}ox
  processes.
\newblock {\em Journal of Machine Learning Research}, 19(67):1--34.

\bibitem[\protect\astroncite{Fernandez et~al.}{2016}]{Fernandez2016GPSurvival}
Fernandez, T., Rivera, N., and Teh, Y.~W. (2016).
\newblock Gaussian processes for survival analysis.
\newblock In {\em NIPS}, pages 5021--5029.

\bibitem[\protect\astroncite{Flaxman et~al.}{2015}]{Flaxman2016CPKronecker}
Flaxman, S., Wilson, A., Neill, D., Nickisch, H., and Smola, A. (2015).
\newblock Fast {K}ronecker inference in {G}aussian processes with
  non-{G}aussian likelihoods.
\newblock In {\em ICML}, pages 607--616.

\bibitem[\protect\astroncite{Flegal et~al.}{2017}]{Flegal2017mcmcse}
Flegal, J.~M., Hughes, J., Vats, D., and Dai, N. (2017).
\newblock {\texttt{mcmcse}}: {M}onte {C}arlo standard errors for {MCMC}.
\newblock \url{https://cran.r-project.org/web/packages/mcmcse/}.

\bibitem[\protect\astroncite{Genz}{1992}]{Genz1992numericalcomputation}
Genz, A. (1992).
\newblock Numerical computation of multivariate normal probabilities.
\newblock {\em Journal of Computational and Graphical Statistics}, 1:141--150.

\bibitem[\protect\astroncite{Gunter et~al.}{2014}]{Gunter2014CPSigmoid}
Gunter, T., Lloyd, C.~M., Osborne, M.~A., and Roberts, S.~J. (2014).
\newblock Efficient {B}ayesian nonparametric modelling of structured point
  processes.
\newblock In {\em UAI}, pages 310--319.

\bibitem[\protect\astroncite{John and Hensman}{2018}]{ST2018CPs}
John, S.~T. and Hensman, J. (2018).
\newblock Large-scale {C}ox process inference using variational {F}ourier
  features.
\newblock In {\em ICML}, pages 2362--2370.

\bibitem[\protect\astroncite{Jones et~al.}{1998}]{Jones1998EGO}
Jones, D.~R., Schonlau, M., and Welch, W.~J. (1998).
\newblock Efficient global optimization of expensive black-box functions.
\newblock {\em Journal of Global Optimization}, 13(4):455--492.

\bibitem[\protect\astroncite{Kingman}{1992}]{Kingman1992PoissonProcess}
Kingman, J. (1992).
\newblock {\em Poisson Processes}.
\newblock Oxford Studies in Probability. Clarendon Press, New York.

\bibitem[\protect\astroncite{Kozachenko et~al.}{2016}]{Kozachenko2016SimCPs}
Kozachenko, Y., Pogorilyak, O., Rozora, I., and Tegza, A. (2016).
\newblock Simulation of {C}ox random processes.
\newblock In {\em Simulation of Stochastic Processes with Given Accuracy and
  Reliability}, pages 251--304. Elsevier, Amsterdam.

\bibitem[\protect\astroncite{Lasko}{2014}]{Lasko2014GPGammaProcesses}
Lasko, T.~A. (2014).
\newblock Efficient inference of {G}aussian-process-modulated renewal processes
  with application to medical event data.
\newblock In {\em UAI}, pages 469--476.

\bibitem[\protect\astroncite{Lloyd et~al.}{2015}]{Lloyd2015CPs}
Lloyd, C.~M., Gunter, T., Osborne, M.~A., and Roberts, S.~J. (2015).
\newblock Variational inference for {G}aussian process modulated {P}oisson
  processes.
\newblock In {\em ICML}, pages 1814--1822.

\bibitem[\protect\astroncite{L\'opez-Lopera}{2018}]{LopezLopera2018LineqGPR}
L\'opez-Lopera, A.~F. (2018).
\newblock {\texttt{lineqGPR}}: Gaussian process regression models with linear
  inequality constraints.
\newblock \url{https://cran.r-project.org/web/packages/lineqGPR/}.

\bibitem[\protect\astroncite{L\'opez-Lopera
  et~al.}{2018}]{LopezLopera2017FiniteGPlinear}
L\'opez-Lopera, A.~F., Bachoc, F., Durrande, N., and Roustant, O. (2018).
\newblock Finite-dimensional {G}aussian approximation with linear inequality
  constraints.
\newblock {\em SIAM/ASA Journal on Uncertainty Quantification},
  6(3):1224--1255.

\bibitem[\protect\astroncite{Maatouk and Bay}{2017}]{Maatouk2017GPineqconst}
Maatouk, H. and Bay, X. (2017).
\newblock Gaussian process emulators for computer experiments with inequality
  constraints.
\newblock {\em Mathematical Geosciences}, 49(5):557--582.

\bibitem[\protect\astroncite{M{\o}ller et~al.}{2001}]{Moller2001LogGaussianCPs}
M{\o}ller, J., Syversveen, A.~R., and Waagepetersen, R.~P. (2001).
\newblock Log {G}aussian {C}ox processes.
\newblock {\em Scandinavian Journal of Statistics}, 25(3):451--482.

\bibitem[\protect\astroncite{M{\o}ller and
  Waagepetersen}{2004}]{Moller2004SpatialPointProcess}
M{\o}ller, J. and Waagepetersen, R.~P. (2004).
\newblock {\em Statistical Inference and Simulation for Spatial Point
  Processes}.
\newblock Chapman \& Hall/CRC Monographs on Statistics and Applied Probability.
  CRC Press, Boca Raton, FL.

\bibitem[\protect\astroncite{Murphy}{2012}]{Murphy2012ML}
Murphy, K.~P. (2012).
\newblock {\em Machine Learning: A Probabilistic Perspective (Adaptive
  Computation And Machine Learning)}.
\newblock The MIT Press, Cambridge.

\bibitem[\protect\astroncite{Pakman and Paninski}{2014}]{Pakman2014Hamiltonian}
Pakman, A. and Paninski, L. (2014).
\newblock Exact {Hamiltonian Monte Carlo} for truncated multivariate
  {G}aussians.
\newblock {\em Journal of Computational and Graphical Statistics},
  23(2):518--542.

\bibitem[\protect\astroncite{Rasmussen and Williams}{2005}]{Rasmussen2005GP}
Rasmussen, C.~E. and Williams, C. K.~I. (2005).
\newblock {\em Gaussian Processes for Machine Learning (Adaptive Computation
  and Machine Learning)}.
\newblock MIT Press, Cambridge.

\bibitem[\protect\astroncite{Riihim\"{a}ki and
  Vehtari}{2010}]{Riihimaki2010GPwithMonotonicity}
Riihim\"{a}ki, J. and Vehtari, A. (2010).
\newblock {G}aussian processes with monotonicity information.
\newblock In {\em AISTATS}, pages 645--652.

\bibitem[\protect\astroncite{Ripley}{1977}]{Ripley1977SpatialPatterns}
Ripley, B.~D. (1977).
\newblock Modelling spatial patterns.
\newblock {\em Journal of the Royal Statistical Society: Series B},
  39(2):172--212.

\bibitem[\protect\astroncite{Serfozo}{2009}]{Serfozo2009RenewalProcesses}
Serfozo, R. (2009).
\newblock {\em Renewal and Regenerative Processes}, pages 99--167.
\newblock Springer, New York.

\bibitem[\protect\astroncite{Sleeper and
  Harrington}{1990}]{Sleeper1990CPSplines}
Sleeper, L.~A. and Harrington, D.~P. (1990).
\newblock Regression splines in the {Cox} model with application to covariate
  effects in liver disease.
\newblock {\em Journal of the American Statistical Association},
  85(412):941--949.

\bibitem[\protect\astroncite{Teh and Rao}{2011}]{Yee2011GPRPs}
Teh, Y.~W. and Rao, V. (2011).
\newblock Gaussian process modulated renewal processes.
\newblock In {\em NIPS}, pages 2474--2482.

\bibitem[\protect\astroncite{Yannaros}{1988}]{Yannaros1988GamaProcesses}
Yannaros, N. (1988).
\newblock On {C}ox processes and {G}amma renewal processes.
\newblock {\em Journal of Applied Probability}, 25(2):423--427.

\bibitem[\protect\astroncite{Yannaros}{1994}]{Yannaros1994WeibullProcess}
Yannaros, N. (1994).
\newblock Weibull renewal processes.
\newblock {\em Annals of the Institute of Statistical Mathematics},
  46(4):641--648.

\end{thebibliography}

%
%
%
%
%

\end{document}